\documentclass[11pt, a4paper, twocolumn, logo, copyright]{eai}
\usepackage{shlab}

\usepackage[authoryear, sort&compress, round]{natbib}
\usepackage{mdframed}

\usepackage{dblfloatfix}
\usepackage{amsmath,amsfonts,bm}
\usepackage{multirow}
\usepackage{subcaption}
\usepackage{wrapfig}

\def\eqref#1{equation~\ref{#1}}

\def\1{\bm{1}}

\DeclareMathAlphabet{\mathsfit}{\encodingdefault}{\sfdefault}{m}{sl}
\SetMathAlphabet{\mathsfit}{bold}{\encodingdefault}{\sfdefault}{bx}{n}

\usepackage{listings}
\usepackage{hyperref}
\usepackage{url}
\usepackage{graphicx}
\usepackage{amsmath}
\usepackage[table]{xcolor}
\usepackage{mathrsfs}
\usepackage{etoolbox}
\usepackage{cleveref}
\usepackage{tcolorbox}
\usepackage{colortbl}
\usepackage{booktabs}
\usepackage{arydshln}
\usepackage{amsfonts}
\usepackage{nicefrac}
\usepackage{microtype}
\usepackage{caption}
\usepackage{subcaption}
\usepackage{algorithm}
\usepackage{wrapfig}
\usepackage{multirow}
\usepackage{lipsum}
\usepackage{makecell}
\usepackage{pifont}
\usepackage{algpseudocode}
\usepackage{algorithmicx}
\usepackage[utf8]{inputenc}
\usepackage[T1]{fontenc}
\usepackage{placeins}
\usepackage{enumitem}
\usepackage{kantlipsum}
\usepackage{fontawesome5}
\usepackage{tikz}

\newcommand{\code}[2]{%
  \begin{tabular}[c]{@{}c@{}}%
    \includegraphics[height=1.3em,width=2.60em,keepaspectratio]{#1}\\[-16pt]%
    {\scriptsize #2}%
  \end{tabular}%
}

\newcommand{\Envelope}{%
  \tikz[
    baseline=-0.5ex,
    x=1.15ex,
    y=0.78ex,
    line width=0.32pt
  ]{%
    \draw[draw=black,fill=white] (0,0) rectangle (1,1);
    \draw[draw=black] (0,1) -- (0.5,0.42) -- (1,1);
  }%
}

\setlist[itemize]{noitemsep, topsep=0pt}

\newlength\savewidth

\definecolor{baselinecolor}{HTML}{d6eaf8}

\definecolor{mygray}{gray}{0.4}

\AtBeginEnvironment{tcolorbox}{\tiny}

\newcount\Comments
\usepackage{color}
\definecolor{darkred}{rgb}{0.9,0,0}
\definecolor{darkgreen}{rgb}{0,0.5,0}
\definecolor{darkblue}{rgb}{0,0,0.7}
\definecolor{purple}{rgb}{.6,0,.6}
\definecolor{orange}{rgb}{1.0,0.64,0}
\definecolor{mgGreen1}{RGB}{238,248,233}
\definecolor{mgGreen2}{RGB}{220,239,211}
\definecolor{mgGreen3}{RGB}{199,228,188}
\definecolor{mgGreen4}{RGB}{177,216,160}
\definecolor{mgGreen5}{RGB}{151,201,130}
\definecolor{supcolor}{rgb}{0.21,0.49,0.74}
\newcommand{\kibitz}[2]{%
  \ifnum\Comments=1\textcolor{#1}{#2}\fi
}

\makeatletter
\patchcmd{\@maketitle}
  {\raggedright}
  {\centering}
  {}{}
\makeatother

\definecolor{cvprblue}{rgb}{0.91,0.96,0.97}
\definecolor{defgreen1}{rgb}{0.910,0.953,0.855}
\definecolor{defgreen2}{rgb}{0.82,0.902,0.710}
\definecolor{defgreen3}{rgb}{0.713,0.903,0.648}
\definecolor{defgreen4}{rgb}{0.725,0.855,0.561}
\definecolor{defyellow}{rgb}{1,0.983,0.717}
\definecolor{orangedot}{HTML}{D07F48}
\definecolor{greendot}{HTML}{71B376}
\definecolor{bluedot}{HTML}{6B96A5}
\definecolor{supplcolor}{HTML}{2C77B9}
\definecolor{revisioncolor}{HTML}{1B6E9E}

\definecolor{thmcolor}{RGB}{0,0,150}

\newtheoremstyle{coloredthm}
  {3pt}
  {3pt}
  {\itshape}
  {}
  {\bfseries\color{thmcolor}}
  {.}
  {.5em}
  {}

\theoremstyle{coloredthm}

\title{\centering GradCO: Behavior-Aware Gradient-Conflict Offloading for Text-Based Person Anomaly Search}

\renewcommand{\today}{2026-10-27}

\newcommand{\copyrightext}{}

\makeatletter

\renewcommand{\footnoterule}{%
  \kern -3pt
  \hrule width \columnwidth height 0.4pt
  \kern 2.6pt
}

\renewcommand{\footrule}{}

\def\@author{%
\makebox[\textwidth][c]{%
\begin{tabular}{c}
Zhiheng Fu\textsuperscript{1}~~ 
Zixu Li\textsuperscript{1,\textdagger}~~
Zhiwei Chen\textsuperscript{1}~~
Jiale Huang\textsuperscript{1}~~ 
Fangxu Liu\textsuperscript{1}~~ 
Jinhe Bi\textsuperscript{2}~~
Yupeng Hu\textsuperscript{1,}\textsuperscript{\Envelope}
\\[3pt]
\textsuperscript{1}Shandong University~~~~
\textsuperscript{2}National University of Singapore
\\
\texttt{\footnotesize\color{black!80} 
\{fuzhiheng8,~lizixu.cs,~zivczw,~cunyurelax\}@gmail.com}
\\

\texttt{\footnotesize\color{black!80}
huangjiale.cs@outlook.com,~bijinhe@outlook.com,~huyupeng@sdu.edu.cn
}
\\
\code{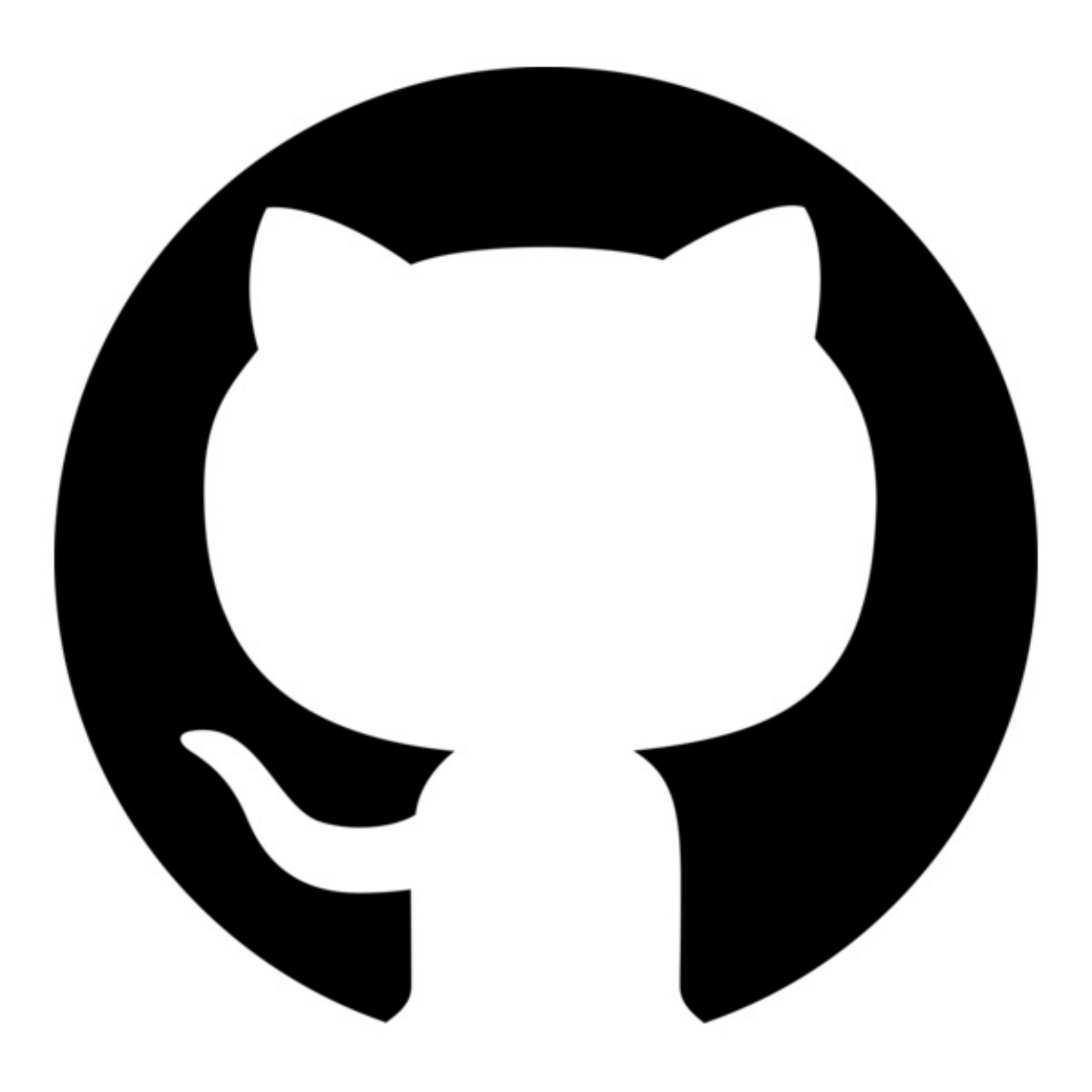}{}\texttt{\footnotesize~Code:~\url{https://github.com/ZhihFu/GradCO_TPAS}}

\end{tabular}%
}
}

\makeatother

\begin{document}

\begin{abstract}

Text-based Person Anomaly Search (TPAS) retrieves a target pedestrian using natural-language descriptions of appearance and normal or anomalous behavior. Unlike conventional text-based person retrieval, TPAS must preserve identity cues while distinguishing fine-grained behaviors. It presents two challenges: same-identity, different-behavior hard negatives cause gradient conflict during granular alignment, while strong appearance cues can obscure weak behavioral evidence during aggregation. We propose \textbf{GradCO} (behavior-aware \textbf{Grad}ient-\textbf{C}onflict \textbf{O}ffloading), comprising three complementary components. Behavior-Aware Mixture-of-Experts (B-MoE) forms appearance, normal-behavior, and anomalous-behavior streams whose roles are learned from retrieval objectives without expert-specific semantic labels. Conflict Offload Learning (COL) uses pair-conditioned offload anchors in similarity computation to reduce the conflicting-gradient burden on the main text representations. Hierarchical Intent Alignment (HIA) aggregates expert similarities and features using query-conditioned weights. Experiments on five TPAS/TIPR benchmarks show GradCO significantly surpasses SOTA methods and validate our design.

\end{abstract}

\maketitle

\begingroup
\renewcommand{\thefootnote}{}
\footnotetext{%
  \parbox[t]{\linewidth}{%
    \textdagger\,Project Leader: Zixu Li.\\
    \Envelope\,Corresponding Author: Yupeng Hu.
  }%
}
\endgroup

\section{Introduction}

\begin{figure}[t]
  \centering
   \includegraphics[width=\linewidth]{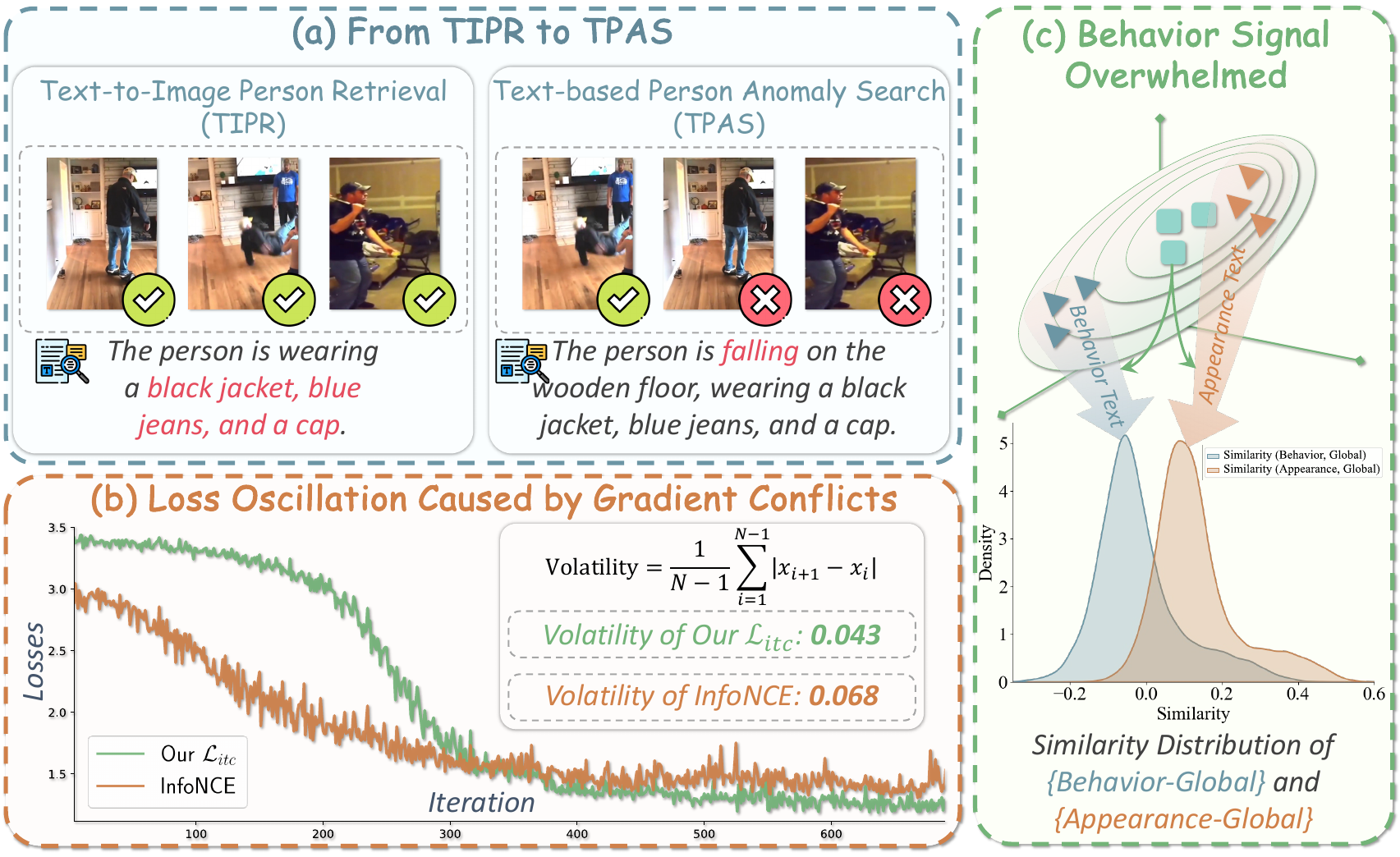}
   \caption{(a)~illustrates examples of TIPR and TPAS tasks. (b)~displays the loss value curves for the InfoNCE loss and our COL-enhanced $\mathcal{L}_{itc}$ (detailed in Sec.~\ref{subsubsec:Conflict Offload Learning} and~\ref{subsubsec: Hierarchical Intent Alignment Network}) on the PAB dataset for TPAS, illustrating the loss oscillation phenomenon caused by gradient conflict. (c)~illustrates that within global features, the feature strength of the appearance signal is typically significantly greater than that of the behavior signal, which readily leads to the intent mismatch phenomenon.}
   \label{fig:intro}
\end{figure}

Text-to-Image Person Retrieval (TIPR) is a key technology in the field of multimodel learning~\citep{Cala,msinet,REFINE,PLOT,CAN,aggarwal2020,GLFN,TIPCB,IRRA}. Recent multimodal large language models further address data selection and modality balance~\citep{Bi2025PRISMSI,song3,bi2025llava,aniri2026opdvvisualonpolicyselfdistillation}. As illustrated on the left side of Figure~\ref{fig:intro}~(a), this technique allows users to locate target pedestrians using natural language descriptions, offering greater flexibility and intuitiveness compared to traditional image-based ReID. However, existing person retrieval research has largely concentrated on matching the static appearance of individuals, such as ``a man wearing a red T-shirt,'' while often overlooking the comprehension of dynamic behaviors, particularly abnormal behaviors that are crucial in safety-critical scenarios \citep{Intelligentsecurity,song6,bialas2010,theodoridis2012,yuan2021,EgoAction,komninos2014}

To bridge the behavioral blind spot in ReID and the identity neglect in traditional visual anomaly detection \citep{ssmtl++,DSVADS,SeS-GCN,normalizing,MoCoDAD}, Yang et al.~\citep{CMP2025} introduced the Text-based Person Anomaly Search (TPAS) task~\citep{CMP2025,LightAIR,xie2026bridging}. As shown in right side of Figure~\ref{fig:intro}~(a), TPAS aims to retrieve a specific pedestrian from a large-scale image database using a natural language query describing both appearance and behavior (normal or abnormal), e.g., ``the man in the blue jacket who is falling down''~\citep{STABLE}.

TPAS requires both fine-grained appearance discrimination from person re-identification (ReID) and subtle behavior perception from anomaly detection, posing a fundamental invariance–sensitivity conflict. 
Specifically, the model must learn appearance features invariant to behavioral changes for identity recognition, while remaining sensitive to postural variations for behavior recognition. A natural solution is to disentangle the feature learning of ``\textit{invariant}'' appearance and ``\textit{sensitive}'' behavior, for example, by using independent identity and behavior expert streams. However, this seemingly reasonable disentanglement scheme, under the unique data distribution of the TPAS task, inevitably induces the following two deeper technical challenges.

\textbf{C1: Gradient Conflict in Granular Alignment}. We observe that hard negative samples in the TPAS task (e.g., ``the same man in a blue jacket who is standing'') are designed to share identity with the positive sample (e.g., ``the man in the blue jacket who is falling down''). A conventional contrastive objective therefore receives competing signals in an appearance-sensitive stream: identity similarity favors preserving the pair's proximity, whereas behavior mismatch favors separating it. This conflict is induced by the joint retrieval objective and does not imply that the corresponding expert stream is assigned an explicit semantic label. As illustrated in Figure~\ref{fig:intro}(b), the COL-enhanced objective exhibits smaller loss fluctuations than standard InfoNCE on PAB, indicating improved optimization stability rather than directly proving semantic disentanglement.

\textbf{C2: Aggregation Distortion of the Alignment Paradigm}. Furthermore, another traditional ``Aggregate-then-Align'' paradigm is also incompatible with the disentanglement scheme. As illustrated in Figure~\ref{fig:intro}(c), if the identity and behavior features are forcibly aggregated (e.g., via concatenation or summation) into a single global vector, the feature strength of the appearance signal can be significantly greater than that of the behavior signal. This aggregation can cause the strong identity (appearance) signal to overwhelm the weak yet critical behavior signal. Moreover, this context-agnostic global feature cannot respond to the dynamic intent of the text (for instance, when querying ``falling down,'' the focus should be primarily on behavior features), leading to an ``\textit{intent mismatch}.''

To address these challenges, we propose \textbf{GradCO} (\textit{Behavior-Aware Gradient-Conflict Offloading}), a TPAS framework that combines three complementary designs.  (a) \textit{Behavior-Aware Mixture-of-Experts (B-MoE)} forms appearance, normal-behavior, and abnormal-behavior streams, whose relative roles emerge from retrieval training without expert-specific semantic labels. (b) \textit{Conflict Offload Learning (COL)} introduces pair-conditioned offload anchors on the text side to redirect conflict-bearing gradients from hard negatives away from the main text representations. (c) \textit{Hierarchical Intent Alignment (HIA)} uses text-conditioned weights to aggregate expert similarities and features, mitigating signal domination and intent mismatch.

The main contributions are summarized as:
\begin{itemize}[leftmargin=8pt]
    \item We identify and analyze two TPAS-specific challenges: gradient conflict from same-identity, different-behavior hard negatives and query-agnostic aggregation distortion.
    \item We propose \textbf{GradCO}, which integrates B-MoE, COL, and HIA for structured expert representation, gradient offloading, and intent-aware aggregation.
    \item Experiments on five TPAS/TIPR benchmarks demonstrate improved retrieval, corruption robustness, and out-of-distribution generalization, supported by ablation and visualization results.
\end{itemize}

\section{Related Work}

\paragraph{Text-based Person Anomaly Search.}
Text-to-Image Person Retrieval (TIPR) retrieves pedestrians from natural-language descriptions~\citep{TIPCB,INTENT,NAFS,song1,Cala,SAF,song2,MIA,IRBTD,PFAN++,ENCODER,msinet,song13}. Existing work uses cross-modal attention~\citep{PLOT,LGUR,IVT,CDCP,REFINE,ReTrack,CAN}, auxiliary cues~\citep{PGMA,aggarwal2020,TEMA}, task-specific objectives~\citep{TIMAM,A-GANet,HGAN}, and vision-language pretraining or large language models~\citep{bi2026the,RaSa,song4,aniri2026fabrimaeitrustmyself,han2021text,song5,bi2026reflectrllearninggoldennegative,IRRA,IVT,CFine,bi2026echorl,clipreid} to align images and text~\citep{ConeSep,Air-know,OFFSET,R3}. In parallel, Video Anomaly Detection (VAD) identifies deviations from normal patterns through latent-, skeleton-, and reconstruction-based paradigms~\citep{SeS-GCN,Dlow,ADMHP,DAflow,Action,DSVADS,MemAE,ssmtl++,AED2017}. TPAS unifies these settings by requiring retrieval conditioned on both pedestrian appearance and normal or anomalous behavior~\citep{CMP2025}. Yet TIPR and VAD methods generally optimize identity matching and behavior recognition separately, leaving their joint alignment unresolved. GradCO addresses this gap by learning a query-responsive joint representation that preserves identity-consistent appearance cues while retaining behavior-sensitive evidence for TPAS retrieval.

\paragraph{Mixtures of Experts.}
Mixture-of-Experts (MoE) architectures route inputs to multiple experts and have been applied to large-model optimization, multitask learning, transfer learning, and ReID~\citep{Adaptive,Switch,Adamv-moe,ST-moe,PRMEM,MPC,li2026lora,META,DaE}. ReID MoE methods mainly model modality variation, biometric-factor fusion, or domain generalization~\citep{MFRNet,Hamobe,DaE}. GradCO extends MoE to this setting by redirecting conflicting optimization signals during granular alignment and making expert aggregation responsive to text intent, so identity consistency and behavior sensitivity can coexist under TPAS hard negatives.

\section{Methodology}

\begin{figure*}[ht!]
  \centering
   \includegraphics[width=\linewidth]{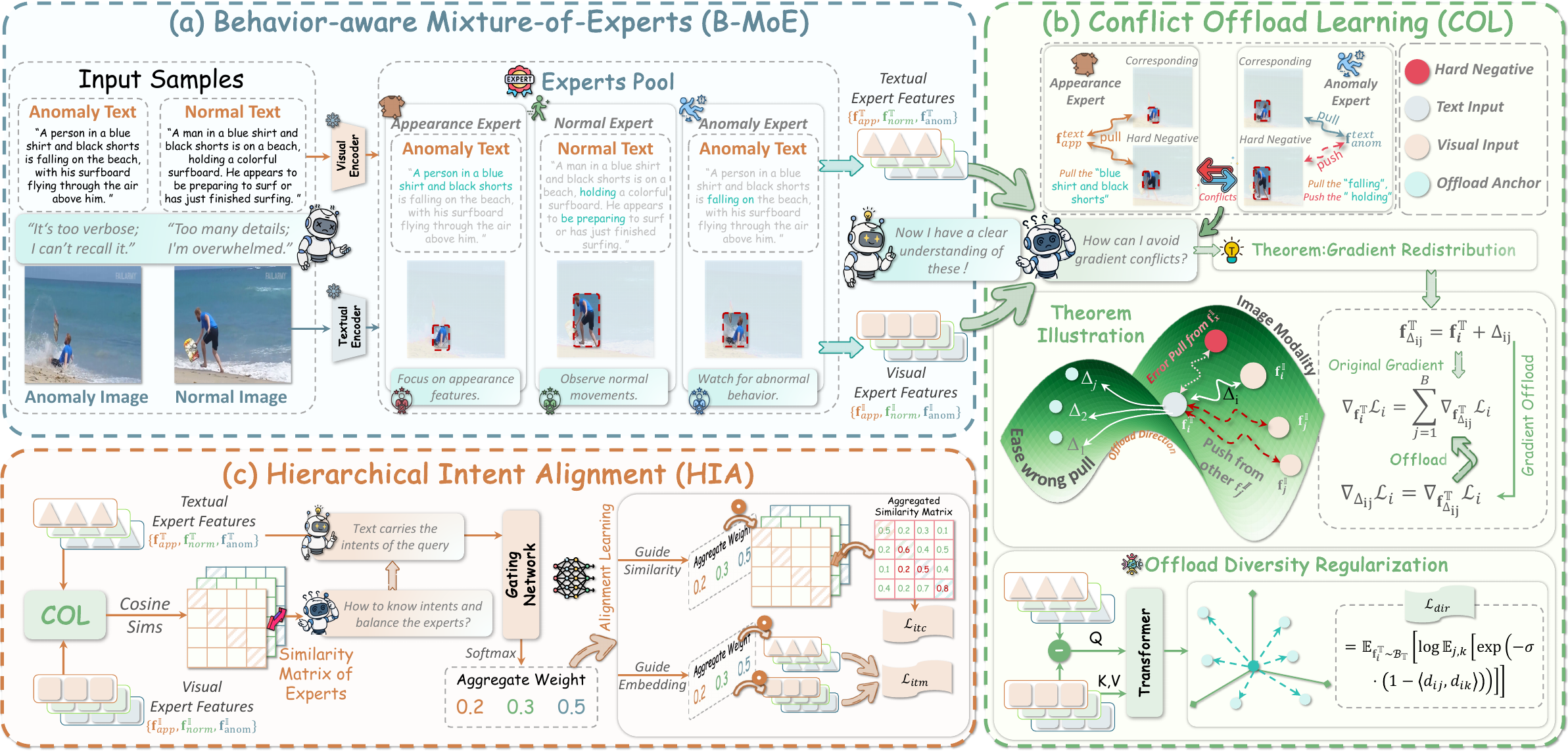}
   \caption{Overview of GradCO. B-MoE learns three input-dependent streams whose names denote post-training routing preferences rather than supervised semantic labels. COL predicts pair-conditioned offload anchors for expert-pair similarities, while HIA aggregates expert similarities and features with query-conditioned weights.}
   \label{fig:framework}
\end{figure*}

GradCO organizes coupled appearance, normal-behavior, and anomal-behavior semantics into parallel expert streams, while offload anchors mitigate alignment-related gradient conflicts. As shown in Figure~\ref{fig:framework}, GradCO consists of three jointly trained components: (a) \textit{Behavior-Aware Mixture-of-Experts (B-MoE, detailed in Sec.~\ref{subsubsec:behavior-Aware Mixture-of-Experts})}; (b) \textit{Conflict Offload Learning (COL, detailed in Sec.~\ref{subsubsec:Conflict Offload Learning})}; and (c) \textit{Hierarchical Intent Alignment (HIA, detailed in Sec.~\ref{subsubsec: Hierarchical Intent Alignment Network})}. We first define the TPAS task and then describe each component.

\subsection{Problem Formulation}
\label{Problem Formulation}
The TPAS task is defined as follows: given a textual query describing the appearance of a pedestrian and a specific normal or abnormal behavior, the objective is to retrieve images from a large scale database where the pedestrian matches both the identity and the described behavior. Formally, given a set of image text pairs $\mathcal{T}=\{(x_\mathbb{T},x_\mathbb{I})_{n}\}_{n=1}^{N}$ (where $x_\mathbb{T}$ and $x_\mathbb{I}$ denote the text and image, respectively, and $N$ is the total number of pairs), the objective is to learn a text encoder $\Phi_\mathbb{T}$ and an image encoder $\Phi_\mathbb{I}$ that map the corresponding text and image into a unified embedding space, such that $\Phi_\mathbb{T}(x_\mathbb{T})\rightarrow \Phi_\mathbb{I}(x_\mathbb{I})$.

\subsection{Behavior-Aware Mixture-of-Experts (B-MoE)}
\label{subsubsec:behavior-Aware Mixture-of-Experts}
To model appearance and behavior cues with different invariance and sensitivity requirements, we design the \textit{Behavior-Aware Mixture-of-Experts (B-MoE)} module. As shown in Figure~\ref{fig:framework}(a), B-MoE constructs three parallel streams for appearance, normal behavior, and abnormal behavior, which provide the representations for subsequent cross-modal alignment.

\paragraph{\textit{Expert Construction.}}
For each modality, every stream contains $K$ sub-experts and a
stream-specific gating network. We use the text-side abnormal-behavior stream
as an example. Given the text encoder output, the sub-experts process the
input from different perspectives and are dynamically combined by
$\mathcal{G}_{ext}$:
\begin{equation}
\mathbf{H}_{anom}^{\mathbb{T}}
=
\sum_{k=1}^{K}
\mathbf{w}_{ext,anom}^{\mathbb{T},k}
\odot
E_{anom}^{\mathbb{T},k}
\bigl(\Phi_{\mathbb{T}}(x_{\mathbb{T}})\bigr),
\end{equation}
where $\mathbf{H}_{anom}^{\mathbb{T}}\in\mathbb{R}^{S\times D}$,
$S$ is the text sequence length, $D$ is the embedding dimension, and
$\mathbf{w}_{ext,anom}^{\mathbb{T},k}$ is the scalar weight of the
$k$-th sub-expert. The appearance and normal-behavior streams use the same
construction to obtain $\mathbf{H}_{app}^{\mathbb{T}}$ and
$\mathbf{H}_{norm}^{\mathbb{T}}$. Although all streams share the encoder
output, their independent experts and gates produce different
input-dependent representations.

\paragraph{\textit{Implicit Specialization.}}
Since three initial representations share the same encoder output, we reweight each according to input context. For the abnormal-behavior
stream, the top-level gate $\mathcal{G}_{top}^{\mathbb{T}}$ produces a scalar
weight $\mathbf{w}_{top,anom}^{\mathbb{T}}\in\mathbb{R}^{1}$, yielding:
\begin{equation}
\mathbf{U}_{anom}^{\mathbb{T}}
=
\mathbf{w}_{top,anom}^{\mathbb{T}}
\odot
\mathbf{H}_{anom}^{\mathbb{T}}.
\label{eq:Top Gating}
\end{equation}

The appearance and normal-behavior streams use their corresponding top-level gates to obtain $\mathbf{U}_{app}^{\mathbb{T}}$ and $\mathbf{U}_{norm}^{\mathbb{T}}$.  No stream-specific semantic labels are used; all sub-experts and gates are jointly optimized by the image-text alignment and matching objectives.

Finally, text projection produces the global text expert features $\mathbf{f}_{app}^{\mathbb{T}}, \mathbf{f}_{norm}^{\mathbb{T}}, \mathbf{f}_{anom}^{\mathbb{T}}\in\mathbb{R}^{D}$. The image branch follows the same procedure with its own modality-specific experts and gates, producing $\mathbf{U}_{app}^{\mathbb{I}}, \mathbf{U}_{norm}^{\mathbb{I}}, \mathbf{U}_{anom}^{\mathbb{I}}\in\mathbb{R}^{C\times D}$, where $C$ is the number of image patches. After image projection, we obtain the global image expert features $\mathbf{f}_{app}^{\mathbb{I}}, \mathbf{f}_{norm}^{\mathbb{I}}, \mathbf{f}_{anom}^{\mathbb{I}}\in\mathbb{R}^{D}$.

\subsection{Conflict Offload Learning (COL)}
\label{subsubsec:Conflict Offload Learning}
To address the gradient conflict problem, we design the
\textit{Conflict Offload Learning (COL)} module based on the expert
features from \textit{B-MoE}. As shown in Figure~\ref{fig:framework}(b),
it introduces pair-conditioned offload anchors for each expert and modifies the gradient path of contrastive learning, thereby providing an offload path for the conflicts caused by ``same-identity, different-behavior'' hard negatives.

\paragraph{\textit{Offload-anchor formulation.}}
Before describing the implementation of COL, we first formulate the
effect of an offload anchor on gradient propagation. For a batch of $B$
paired samples, let $\mathbf{S}$ denote image-text similarity matrix, where
$s_{ij}=\cos(\mathbf{f}^{\mathbb{T}}_i,\mathbf{f}^{\mathbb{I}}_j)$.
The InfoNCE loss is defined as follows,
\begin{equation}
\mathcal{L}_{\mathrm{NCE}}(\mathbf{S})
\!=\!
-\frac{1}{B}\sum_{i=1}^{B}
\log\left(
\frac{\exp(s_{ii}/\tau)}
{\sum_{j=1}^{B}\exp(s_{ij}/\tau)}
\right),
\label{infonce}
\end{equation}
where $\tau$ is the temperature coefficient.

For each sample pair, we introduce an offload anchor $\Delta_{ij}$ and
redefine the similarity as follows,
\begin{equation}
\hat{s}_{ij}
=
\cos(\mathbf{f}^{\mathbb{T}}_i+\Delta_{ij},
\mathbf{f}^{\mathbb{I}}_j).
\end{equation}
Let
$\hat{p}_{ij}=
\exp(\hat{s}_{ij}/\tau)/
\sum_{\ell=1}^{B}\exp(\hat{s}_{i\ell}/\tau)$,
$y_{ij}=\mathbf{1}_{[j=i]}$,
$\mathbf{g}_{ij}=\nabla_{\mathbf{z}_{ij}}\hat{s}_{ij}$ with
$\mathbf{z}_{ij}=\mathbf{f}^{\mathbb{T}}_i+\Delta_{ij}$, and
$\mathbf{J}_{ij}
=\partial\Delta_{ij}/\partial\mathbf{f}^{\mathbb{T}}_i$.
The corresponding gradients are
\begin{equation}
\begin{aligned}
\nabla_{\Delta_{ij}}\mathcal{L}_{i}
&=
\frac{\hat{p}_{ij}-y_{ij}}{\tau}\mathbf{g}_{ij},\\
\nabla_{\mathbf{f}^{\mathbb{T}}_i}\mathcal{L}_{i}
&=
\sum_{j=1}^{B}
(\mathbf{I}+\mathbf{J}_{ij})^{\top}
\nabla_{\Delta_{ij}}\mathcal{L}_{i},
\end{aligned}
\label{gradient_dis}
\end{equation}
where $\mathbf{I}$ is the identity matrix. The first line gives the
gradient through the offload path, while the second accounts for both
the direct path through
$\mathbf{f}^{\mathbb{T}}_i+\Delta_{ij}$ and the dependence of the
predicted anchor on the text feature through $\mathbf{J}_{ij}$.

We focus on text-side gradients because text acts as the dynamic query while image gallery remains fixed at retrieval time. The pair-conditioned anchor absorbs part of conflicting alignment pressure while preserving text feature's semantic content.

\paragraph{\textit{Conflict offloading.}}
For each expert image-text pair $(\mathbf{f}^{\mathbb{T}}_i,\mathbf{f}^{\mathbb{I}}_j)$, COL computes an offload anchor from their feature difference and the image expert tokens. We illustrate the process using the abnormal-behavior expert:
\begin{equation}
\begin{aligned}
\Delta_{\mathrm{anom}}^{ij}
&=
\mathrm{TF}_{\mathrm{anom}}\Big(
Q=
\mathbf{f}_{\mathrm{anom},i}^{\mathbb{T}}
-
\mathbf{f}_{\mathrm{anom},j}^{\mathbb{I}},\\[-2pt]
&\hspace{22mm}
\{K,V\}
=
\mathbf{U}_{\mathrm{anom},j}^{\mathbb{I}}
\Big),
\end{aligned}
\end{equation}
where $\mathrm{TF}_{\mathrm{anom}}$ denotes the Transformer used by the
abnormal-behavior expert. The corresponding offload-based similarity is as follows,
\begin{equation}
s_{\mathrm{anom},ij}
=
\cos\left(
\mathbf{f}_{\mathrm{anom},i}^{\mathbb{T}}
+
\Delta_{\mathrm{anom}}^{ij},
\mathbf{f}_{\mathrm{anom},j}^{\mathbb{I}}
\right).
\end{equation}
The appearance and normal-behavior experts follow the same procedure,
yielding $\mathbf{S}_{\mathrm{app}}$ and $\mathbf{S}_{\mathrm{norm}}$.

\paragraph{\textit{Direction diversity regularization.}} 
To prevent the offload anchors from collapsing into similar directions, we introduce:
\begin{equation}
\begin{aligned}
\mathcal{L}_{\mathrm{dir}}
&=
\mathbb{E}_{\mathbf{f}_{\mathrm{anom},i}^{\mathbb{T}}
\sim\mathcal{B}_{\mathbb{T}}}
\Big[
\log
\mathbb{E}_{j\neq k}
\Big[
\\[-2pt]
&\exp\Big(
-\sigma
\big(
1-\langle d_{ij},d_{ik}\rangle
\big)
\Big)
\Big]
\Big],
\end{aligned}
\label{eq:dir}
\end{equation}
where $\mathcal{B}_{\mathbb{T}}$ denotes the text samples in the batch,
$\sigma$ is a scaling factor, and $d_{ij}$ and $d_{ik}$ are the
normalized forms of
$\Delta_{\mathrm{anom}}^{ij}$ and
$\Delta_{\mathrm{anom}}^{ik}$, respectively. 
$\langle\cdot,\cdot\rangle$ denotes cosine similarity. This regularizer
encourages different pair-conditioned anchors associated with the same
text query to maintain distinct directions.

\subsection{Hierarchical Intent Alignment (HIA)}
\label{subsubsec: Hierarchical Intent Alignment Network}

To address aggregation distortion, we design \textit{Hierarchical Intent Alignment (HIA)}. As shown in
Figure~\ref{fig:framework}(c), HIA predicts query-conditioned weights from
text expert features and uses them to aggregate expert similarities and
features.
\paragraph{\textit{Query-conditioned similarity aggregation.}}
Taking the text side as an example, we feed the three global expert features
from \textit{B-MoE} into the text intent gating network
$\mathcal{G}_{H}^{\mathbb{T}}$. For the $i$-th text query, it produces
$\mathbf{W}_{gate,i}^{\mathbb{T}}\in\mathbb{R}^{3}$:
\begin{equation}
\begin{aligned}
\mathbf{W}_{gate,i}^{\mathbb{T}}
  &=
  \mathrm{Softmax}\!\Bigl(
  \mathcal{G}_{H}^{\mathbb{T}}\bigl(
  \mathrm{Concat}[
  \mathbf{f}_{app,i}^{\mathbb{T}},\\[-2pt]
  &\hspace{22mm}
  \mathbf{f}_{norm,i}^{\mathbb{T}},
  \mathbf{f}_{anom,i}^{\mathbb{T}}]
  \bigr)\Bigr),
\end{aligned}
\label{eq:gating}
\end{equation}
where
$\mathbf{W}_{gate,i}^{\mathbb{T}}
=
[w_{app,i}^{\mathbb{T}},w_{norm,i}^{\mathbb{T}},w_{anom,i}^{\mathbb{T}}]$.
For a text-image pair $(i,j)$, these weights aggregate the
expert similarities as
$S_{\mathrm{final}}^{ij}
=
\sum_{r\in\mathcal{K}}w_{r,i}^{\mathbb{T}}S_{r}^{ij}$, where
$\mathcal{K}=\{app,norm,anom\}$. We then optimize the bidirectional
image-text contrastive loss:
\begin{equation}
\mathcal{L}_{itc}
=
\frac{1}{2}
\left[
\mathcal{L}_{\mathrm{NCE}}(\mathbf{S}_{\mathrm{final}})
+
\mathcal{L}_{\mathrm{NCE}}(\mathbf{S}_{\mathrm{final}}^{\mathsf T})
\right],
\label{eq:itc}
\end{equation}
where $\mathcal{L}_{\mathrm{NCE}}(\cdot)$ is defined in
Eq.~(\ref{infonce}). This loss jointly trains expert streams and HIA
gate, allowing their relative contributions to adapt to each query.

\paragraph{\textit{Intent-aware feature aggregation and matching.}}
Following CMP~\citep{CMP2025}, we further introduce an image-text matching
loss to model semantic combinations. We reuse the weights in
Eq.~(\ref{eq:gating}) to aggregate the text expert features as
$\mathbf{f}_{\mathrm{final},i}^{\mathbb{T}}
=
\sum_{r\in\mathcal{K}}
w_{r,i}^{\mathbb{T}}\mathbf{f}_{r,i}^{\mathbb{T}}$.
For each image-text pair, the same weights are applied to the corresponding
image expert features. The resulting representations are fused by last
six layers of BERT, which act as a cross encoder and produce the matching
logit $s_{itm}$. We optimize the binary cross-entropy loss
$\mathcal{L}_{itm}$ using the label
$y_{itm}\in\{0,1\}$.

Finally, the overall objective is as follows, 
\begin{equation}
\mathbf{\Theta}^{*}
=
\underset{\mathbf{\Theta}}{\arg\min}
\left(
\mathcal{L}_{itc}
+
\mathcal{L}_{itm}
+
\gamma\mathcal{L}_{dir}
\right),
\label{optimization}
\end{equation}
where $\gamma$ weights the direction-diversity regularizer. Following
previous works~\citep{CMP2025,Bi-IRRA}, the coefficients of
$\mathcal{L}_{itc}$ and $\mathcal{L}_{itm}$ are set to $1$. 

\begin{table}[t]
  \centering
  \caption{Performance comparison on PAB dataset. Best results are in \textbf{bold}, and sub-optimal is \underline{underlined}.}
    \resizebox{\linewidth}{!}{%
    \begin{tabular}{cccccc}
 \Xhline{1.5pt}
    \multicolumn{1}{c|}{Method} & \multicolumn{1}{c|}{\#Data} & \multicolumn{1}{c}{R@1} & \multicolumn{1}{c}{R@5} & \multicolumn{1}{c}{R@10} & \multicolumn{1}{c}{mAP} \\
    \hline
    \multicolumn{1}{l|}{CLIP~\textcolor{gray}{(ICML'21)}} & \multicolumn{1}{c|}{-} & \multicolumn{1}{c}{47.57 } & \multicolumn{1}{c}{81.55 } & \multicolumn{1}{c}{89.03 } & \multicolumn{1}{c}{62.73 } \\
    \multicolumn{1}{l|}{X-VLM~\textcolor{gray}{(ICML'22)}} & \multicolumn{1}{c|}{-} & \multicolumn{1}{c}{71.94 } & \multicolumn{1}{c}{97.78 } & \multicolumn{1}{c}{98.99 } & \multicolumn{1}{c}{83.96 } \\
    \multicolumn{1}{l|}{RaSa~\textcolor{gray}{(IJCAI'23)}} & \multicolumn{1}{c|}{-} & \multicolumn{1}{c}{21.74 } & \multicolumn{1}{c}{27.30 } & \multicolumn{1}{c}{27.96 } & \multicolumn{1}{c}{24.35 } \\
    \multicolumn{1}{l|}{APTM~\textcolor{gray}{(MM'23)}} & \multicolumn{1}{c|}{-} & \multicolumn{1}{c}{22.90 } & \multicolumn{1}{c}{45.80 } & \multicolumn{1}{c}{52.38 } & \multicolumn{1}{c}{33.56 } \\
    \multicolumn{1}{l|}{IRRA~\textcolor{gray}{(CVPR'23)}} & \multicolumn{1}{c|}{-} & \multicolumn{1}{c}{30.59 } & \multicolumn{1}{c}{59.61 } & \multicolumn{1}{c}{68.91 } & \multicolumn{1}{c}{44.41 } \\ 
    \multicolumn{1}{l|}{MRA~\textcolor{gray}{(arXiv'25)}} & \multicolumn{1}{c|}{-} & \multicolumn{1}{c}{9.91 } & \multicolumn{1}{c}{23.66 } & \multicolumn{1}{c}{31.45 } & \multicolumn{1}{c}{17.15 } \\    
    \multicolumn{1}{l|}{WoRA~\textcolor{gray}{(WWW'25)}} & \multicolumn{1}{c|}{-} & \multicolumn{1}{c}{22.25 } & \multicolumn{1}{c}{45.91 } & \multicolumn{1}{c}{53.54 } & \multicolumn{1}{c}{33.39 } \\    
    \multicolumn{1}{l|}{CAMeL~\textcolor{gray}{(TIFS'25)}} & \multicolumn{1}{c|}{-} & \multicolumn{1}{c}{24.47 } & \multicolumn{1}{c}{50.00 } & \multicolumn{1}{c}{58.75 } & \multicolumn{1}{c}{36.75 } \\
    \hline
    \multicolumn{1}{l|}{CLIP~\textcolor{gray}{(ICML'21)}} & \multicolumn{1}{c|}{0.1M} & \multicolumn{1}{c}{77.60 } & \multicolumn{1}{c}{98.84 } & \multicolumn{1}{c}{99.75 } & \multicolumn{1}{c}{87.35 } \\
    \multicolumn{1}{l|}{X-VLM~\textcolor{gray}{(ICML'22)}} & \multicolumn{1}{c|}{0.1M} & \multicolumn{1}{c}{81.95 } & \multicolumn{1}{c}{98.84 } & \multicolumn{1}{c}{99.19 } & \multicolumn{1}{c}{89.86 } \\
    \multicolumn{1}{l|}{RaSa~\textcolor{gray}{(IJCAI'23)}} & \multicolumn{1}{c|}{0.1M} & \multicolumn{1}{c}{80.79 } & \multicolumn{1}{c}{98.89 } & \multicolumn{1}{c}{99.65 } & \multicolumn{1}{c}{89.20 } \\
    \multicolumn{1}{l|}{APTM~\textcolor{gray}{(MM'23)}} & \multicolumn{1}{c|}{0.1M} & \multicolumn{1}{c}{72.14 } & \multicolumn{1}{c}{95.30 } & \multicolumn{1}{c}{97.17 } & \multicolumn{1}{c}{82.78 } \\
    \multicolumn{1}{l|}{IRRA~\textcolor{gray}{(CVPR'23)}} & \multicolumn{1}{c|}{0.1M} & \multicolumn{1}{c}{76.39 } & \multicolumn{1}{c}{97.62 } & \multicolumn{1}{c}{99.14 } & \multicolumn{1}{c}{86.33 } \\
    \multicolumn{1}{l|}{MRA~\textcolor{gray}{(arXiv'25)}} & \multicolumn{1}{c|}{0.1M} & \multicolumn{1}{c}{70.53 } & \multicolumn{1}{c}{94.69 } & \multicolumn{1}{c}{97.47 } & \multicolumn{1}{c}{81.59 } \\ 
    \multicolumn{1}{l|}{WoRA~\textcolor{gray}{(WWW'25)}} & \multicolumn{1}{c|}{0.1M} & \multicolumn{1}{c}{74.47 } & \multicolumn{1}{c}{96.82 } & \multicolumn{1}{c}{98.48 } & \multicolumn{1}{c}{84.60 } \\
    \multicolumn{1}{l|}{CAMeL~\textcolor{gray}{(TIFS'25)}} & \multicolumn{1}{c|}{0.1M} & \multicolumn{1}{c}{74.30 } & \multicolumn{1}{c}{96.79 } & \multicolumn{1}{c}{98.84 } & \multicolumn{1}{c}{84.20 } \\    
    \multicolumn{1}{l|}{CMP~\textcolor{gray}{(ICCV'25)}} & \multicolumn{1}{c|}{0.1M} & \multicolumn{1}{c}{83.06 } & \multicolumn{1}{c}{98.89 } & \multicolumn{1}{c}{99.49 } & \multicolumn{1}{c}{90.41 } \\
    \rowcolor[rgb]{0.949, 0.949, 1.0} 
    \multicolumn{1}{l|}{\textbf{GradCO~(Ours)}} & \multicolumn{1}{c|}{0.1M} & \multicolumn{1}{c}{\underline{85.39} } & \multicolumn{1}{c}{\underline{99.34} } & \multicolumn{1}{c}{\underline{99.60} } & \multicolumn{1}{c}{\underline{92.13} } \\
    \hline
    \multicolumn{1}{l|}{CMP~\textcolor{gray}{(ICCV'25)}} & \multicolumn{1}{c|}{1M} &
    \multicolumn{1}{c}{84.93 } & \multicolumn{1}{c}{99.09 } & \multicolumn{1}{c}{\textbf{99.75} } & \multicolumn{1}{c}{91.66 } \\
    \rowcolor[rgb]{0.949, 0.949, 1.0} 
    \multicolumn{1}{l|}{\textbf{GradCO~(Ours)}} & \multicolumn{1}{c|}{1M} & \multicolumn{1}{c}{\textbf{86.60}} & \multicolumn{1}{c}{	\textbf{99.60}	 } & \multicolumn{1}{c}{\textbf{99.75}} & \multicolumn{1}{c}{	\textbf{92.88}}  \\
 \Xhline{1.5pt}
    \end{tabular}%
    }
  \label{tab:pab}%
\end{table}%

\section{Experiments}
In this section, we evaluate GradCO on the target TPAS task and conventional TIPR benchmarks, followed by ablation, sensitivity, and qualitative analyses.

\begin{table*}[ht]
  \centering
  \caption{Performance comparison on the PAB dataset of multi-weather test setting, measured by R@1 and mAP for each weather. Our model is highlighted in light purple, and the overall best results are indicated in \textbf{bold}.}
    \resizebox{\linewidth}{!}{%
    \begin{tabular}{c|cc|cc|cc|cc|cc|cc|cc|cc|cc|cc|cc}
 \Xhline{1.5pt}
    \multirow{2}{*}{Method} & \multicolumn{2}{c|}{Normal} & \multicolumn{2}{c|}{Wind} & \multicolumn{2}{c|}{Rain} & \multicolumn{2}{c|}{Snow} & \multicolumn{2}{c|}{Rain+Snow} & \multicolumn{2}{c|}{Dark} & \multicolumn{2}{c|}{Dark+Wind} & \multicolumn{2}{c|}{Dark+Rain} & \multicolumn{2}{c|}{Dark+Snow} & \multicolumn{2}{c|}{Over-exposure} & \multicolumn{2}{c}{Mean} \\
    \multicolumn{1}{c|}{} & \multicolumn{1}{c}{R@1} & \multicolumn{1}{c|}{mAP} & \multicolumn{1}{c}{R@1} & \multicolumn{1}{c|}{mAP} & \multicolumn{1}{c}{R@1} & \multicolumn{1}{c|}{mAP} & \multicolumn{1}{c}{R@1} & \multicolumn{1}{c|}{mAP} & \multicolumn{1}{c}{R@1} & \multicolumn{1}{c|}{mAP} & \multicolumn{1}{c}{R@1} & \multicolumn{1}{c|}{mAP} & \multicolumn{1}{c}{R@1} & \multicolumn{1}{c|}{mAP} & \multicolumn{1}{c}{R@1} & \multicolumn{1}{c|}{mAP} & \multicolumn{1}{c}{R@1} & \multicolumn{1}{c|}{mAP} & \multicolumn{1}{c}{R@1} & \multicolumn{1}{c|}{mAP} & \multicolumn{1}{c}{R@1} & \multicolumn{1}{c}{mAP} \\
    \hline
    \multicolumn{1}{l|}{X-VLM~\textcolor{gray}{(ICML'22)}} & 83.47  & 90.94  & 79.02  & 88.10  & 54.40  & 67.40  & 59.10  & 72.34  & 49.95  & 62.09  & 79.58  & 88.24  & 75.53  & 85.47  & 34.93  & 45.98  & 50.25  & 63.32  & 74.87  & 84.82  & 64.11  & 74.87  \\
    \multicolumn{1}{l|}{CMP~\textcolor{gray}{(ICCV'25)}} & 84.93  & 91.66  & 81.24  & 89.34  & 60.06  & 72.53  & 63.40  & 75.74  & 54.85  & 67.31  & 80.89  & 89.00  & 77.20  & 86.55  & 39.03  & 50.58  & 53.49  & 66.12  & 76.09  & 85.56  & 67.12  & 77.44  \\
    \hline
    \rowcolor[rgb]{0.949, 0.949, 1.0} 
    \multicolumn{1}{l|}{\textbf{GradCO~(Ours)}}  & \textbf{86.60}  & \textbf{92.88}  & \textbf{83.69}  & \textbf{91.25}  & \textbf{68.00}  & \textbf{78.85}  & \textbf{69.55}  & \textbf{80.68}  & \textbf{61.77}  & \textbf{69.20}  & \textbf{83.24}  & \textbf{90.08}  & \textbf{82.08}  & \textbf{89.33}  & \textbf{47.38}  & \textbf{55.22}  & \textbf{60.31}  & \textbf{70.21}  & \textbf{79.32}  & \textbf{88.95}  & \textbf{72.20}  & \textbf{80.67} \\
    \Xhline{1.5pt}
    \end{tabular}%
    }
  \label{tab:multiweather}%
\end{table*}%

\subsection{Experimental Settings}
\label{sec:experimental_settings}

\subsubsection{Datasets.}
For TPAS, we use PAB~\citep{CMP2025} as the in-distribution benchmark, Multi-Weather to evaluate robustness to weather corruptions, and UCC as the out-of-distribution (OOD) test set. For TIPR, we evaluate on CUHK-PEDES~\citep{CUHK-PEDES}, ICFG-PEDES~\citep{ICFG-PEDES}, RSTPReID~\citep{RSTPReID}, and UFineBench~\citep{UFineBench}, which includes the UFine6926 and UFine3C settings.

\noindent \textbf{PAB.}
Pedestrian Anomaly Behavior (PAB)~\cite{CMP2025} is a benchmark for
text-based person anomaly search, which retrieves pedestrians engaged
in normal or anomalous activities using natural-language descriptions.
It contains 1,015,583 image-text pairs, including 1,013,605 synthetic
training pairs generated with diffusion models using descriptions from
OOPS! as prompts, and 1,978 real-world test pairs from OOPS!. PAB covers
1,600 anomalous action types (e.g., falling and being hit) and 1,000
normal action types (e.g., running and playing soccer). Its frame-level
descriptions cover appearance, actions, and scenes, with an average
length of 50.3 words; the training set uses an anomalous-to-normal ratio
of 3:2. To construct the test set, OpenPose and ResNet-50 are used for
filtering, followed by professional manual verification. Qwen2-VL
generates the test descriptions, which are then checked through human
quality control.

\noindent \textbf{Multi-Weather Setting of PAB.}
Multi-Weather is an evaluation protocol applied to the PAB test set to
measure retrieval robustness under weather and lighting corruptions.
It includes ten conditions: Normal, Wind, Rain, Snow, Rain+Snow, Dark,
Dark+Wind, Dark+Rain, Dark+Snow, and Over-exposure. We report results
for each condition and their arithmetic mean.

\noindent \textbf{UCC.}
UCC is a real-world test set for evaluating out-of-distribution (OOD)
generalization from PAB to an unseen data source. It is derived from
UCF-Crime and contains 5,320 image-text pairs. The images are selected
keyframes from 13 normal and anomalous video categories, and Qwen2-VL
generates the corresponding textual descriptions.

\begin{table}[t]
  \centering
  \caption{Comparisons with existing methods in OOD setting. Best results are indicated in \textbf{bold}, and the overall second-best results are \underline{underlined}.}
    \resizebox{0.9\linewidth}{!}{%
    \begin{tabular}{c|cccc}
 \Xhline{1.5pt}
    Method & \multicolumn{1}{c}{R@1} & \multicolumn{1}{c}{R@5} & \multicolumn{1}{c}{R@10} & \multicolumn{1}{c}{mAP} \\
    \hline
    \multicolumn{1}{l|}{CLIP~\textcolor{gray}{(ICML'21)}}  & 51.60  & 68.31  & 76.43  & 43.05  \\
    \multicolumn{1}{l|}{X-VL~\textcolor{gray}{(ICML'22)}} & 52.33  & 66.73  & 72.54  & 40.87  \\
    \multicolumn{1}{l|}{APTM~\textcolor{gray}{(MM'23)}}& 27.86  & 40.41  & 46.77  & 22.61  \\
    \multicolumn{1}{l|}{IRRA~\textcolor{gray}{(CVPR'23)}}   & 40.28  & 57.24  & 65.98  & 33.53  \\
    \multicolumn{1}{l|}{RaSa~\textcolor{gray}{(IJCAI'23)}}  & 54.12  & 70.32  & 75.96  & 39.71  \\
    \multicolumn{1}{l|}{CMP~\textcolor{gray}{(ICCV'25)}}  & 54.12  & 71.07  & 77.90  & 43.13  \\
    \multicolumn{1}{l|}{CMP(1M)~\textcolor{gray}{(ICCV'25)}} & 55.23  & 71.67  & 77.99  & 44.35  \\
    \hline
    \rowcolor[rgb]{0.949, 0.949, 1.0} 
    \multicolumn{1}{l|}{\textbf{GradCO}}  & \underline{61.85} & \underline{78.31} & \underline{86.09} & \underline{51.78}  \\
    \rowcolor[rgb]{0.949, 0.949, 1.0} 
    \multicolumn{1}{l|}{\textbf{GradCO~(1M)}} & \textbf{62.53} & \textbf{78.81} & \textbf{86.89} & \textbf{52.01}  \\
 \Xhline{1.5pt}
    \end{tabular}%
    }
    \vspace{-8pt}
  \label{tab:ood}%
\end{table}%

\begin{table}[t]
  \centering
  \caption{Performance comparison on TIPR task datasets. Best results are shown in bold, and the second-best results are \underline{underlined}.}
    \resizebox{\linewidth}{!}{%
    \begin{tabular}{c|c|c|c|cc}
 \Xhline{1.5pt}
    \multirow{2}[1]{*}{Method} & CUHK  & ICFG  & RSTP  & \multicolumn{2}{c}{UFineBench} \\
          & Avg   & Avg   & Avg   & 6926-Avg. & 3C-Avg. \\
     \hline
    \multicolumn{1}{l|}{NAFS~\textcolor{gray}{(ECCV'18)}}  & - & - & - & 76.49  & 58.25  \\
    \multicolumn{1}{l|}{SSAN~\textcolor{gray}{(arXiv'21)}}  & - & - & - & 85.52  & 67.32  \\
    \multicolumn{1}{l|}{CMKA~\textcolor{gray}{(TIP'21)}}  & 70.07  & - & - & - & - \\
    \multicolumn{1}{l|}{LapsCore~\textcolor{gray}{(ICCV'21)}} & 75.60  & - & - & - & - \\
    \multicolumn{1}{l|}{LGUR~\textcolor{gray}{(MM'22)}}  & - & - & - & 81.72  & 65.42  \\    
    \multicolumn{1}{l|}{SAF~\textcolor{gray}{(ICASSP'22)}}   & 78.38  & - & - & - & - \\
    \multicolumn{1}{l|}{TIPCB~\textcolor{gray}{(Neuro'22)}} & 78.85  & - & - & - & - \\
    \multicolumn{1}{l|}{AXM-Net~\textcolor{gray}{(MM'22)}} & 77.24  & - & - & - & - \\
    \multicolumn{1}{l|}{MANet~\textcolor{gray}{(TNNLS'23)}} & 79.14  & 73.00  & - & - & - \\
    \multicolumn{1}{l|}{CFine~\textcolor{gray}{(TIP'23)}} & 82.22  & 73.27  & 68.22  & - & - \\
    \multicolumn{1}{l|}{IRRA~\textcolor{gray}{(CVPR'23)}}  & 85.67  & 76.51  & 76.57  & 90.81  & 68.99  \\
    \multicolumn{1}{l|}{BiLMa~\textcolor{gray}{(ICCV'23)}} & 85.75  & 76.57  & 77.17  & - & - \\
    \multicolumn{1}{l|}{RaSa~\textcolor{gray}{(IJCAI'23)}}  & 87.02  & 76.93  & 81.58  & - & - \\
    \multicolumn{1}{l|}{TBPS-CLIP~\textcolor{gray}{(AAAI'24)}} & 84.69  & 76.95  & 78.08  & - & - \\
    \multicolumn{1}{l|}{CADA-G~\textcolor{gray}{(TMM'24)}} & 85.72  & 75.71  & 77.75  & - & - \\
    \multicolumn{1}{l|}{UMSA~\textcolor{gray}{(AAAI'24)}}  & 85.89  & 77.33  & 79.00  & - & - \\
    \multicolumn{1}{l|}{FSRL~\textcolor{gray}{(ICMR'24)}}  & 86.32  & 77.28  & 77.77  & - & - \\
    \multicolumn{1}{l|}{Propot~\textcolor{gray}{(MM'24)}} & 86.32  & 77.89  & 78.40  & - & - \\
    \multicolumn{1}{l|}{CFAM~\textcolor{gray}{(CVPR'24)}}  & 86.83  & 77.63  & 79.03  & 93.86  & \underline{74.63}  \\
    \multicolumn{1}{l|}{RDE~\textcolor{gray}{(CVPR'24)}}   & 86.73  & 79.17  & 79.73  & - & - \\
    \multicolumn{1}{l|}{Bi-IRRA~\textcolor{gray}{(TPAMI'25)}} & \underline{88.77}  & \underline{79.79}  & \underline{84.17}  & \textbf{95.18}  & - \\
    \hline
    \rowcolor[rgb]{0.949, 0.949, 1.0} 
    \multicolumn{1}{l|}{\textbf{GradCO~(Ours)}}  & \textbf{88.87} & \textbf{79.85} & \textbf{84.73} & \underline{95.13} & \textbf{82.01} \\
    \Xhline{1.5pt}
    \end{tabular}%
    }
  \label{tab:tipr}%
\end{table}%

\begin{table*}[t]
  \centering
  \caption{Comparison of computational efficiency and performance between GradCO and the baseline method CMP. GradCO is highlighted in light purple.}
  \resizebox{\linewidth}{!}{%
    \begin{tabular}{c|c|c|c|c|c|c|c}
    \Xhline{1.5pt}
    Method  & Parameters(M) & GPU Memory(MiB) & Test time(s/sample) & Train Time(s/iteration) & PAB-Avg & MultiWeather-Avg & UCC(OOD)-Avg \\
    \hline
    CMP   & 226.13 & 14575~(bs=22) & 0.0015 & 0.5326~(bs=22) & 94.59 & 67.12 & 68.30  \\
    \rowcolor[rgb]{0.949, 0.949, 1.0} \textbf{GradCO (Ours)}     & 247.36  &  18413~(bs=22)    & 0.0023     & 0.7213~(bs=22)     & \textbf{95.32}     & \textbf{72.19}  &  \textbf{76.08} \\
    \Xhline{1.5pt}
    \end{tabular}%
    }
  \label{tab:efficiency_comparison}%
\end{table*}%

\noindent \textbf{CUHK-PEDES.}
CUHK-PEDES~\cite{CUHK-PEDES} is a text-to-image person retrieval
benchmark containing 40,206 images and 80,440 descriptions from 13,003
identities. Its training set contains 34,054 images, 68,126 descriptions,
and 11,003 identities; the test set contains 3,074 images, 6,156
descriptions, and 1,000 identities.

\noindent \textbf{ICFG-PEDES.}
ICFG-PEDES~\cite{ICFG-PEDES} contains 54,522 images from 4,102
identities, with a text description paired with each image. Its
training set contains 34,674 images from 3,102 identities, and its test
set contains 19,848 images from the remaining 1,000 identities.

\noindent \textbf{RSTPReID.}
RSTPReID~\cite{RSTPReID} contains 20,505 images of 4,101 identities
collected from 15 cameras. Each identity has five images captured by
different cameras, and each image is annotated with two text
descriptions. The dataset is split into 3,701 training, 200 validation,
and 200 test identities.

\noindent \textbf{UFineBench.}
UFineBench~\cite{UFineBench} contains UFine6926 and the UFine3C
evaluation set. UFine6926 has 26,206 images and 52,412 descriptions
from 6,926 identities. Its descriptions are ultra-fine-grained, with
an average length of 80.8 words. The training set contains 18,577
images and 37,154 descriptions from 4,926 identities, while the test
set contains 7,629 images and 15,258 descriptions from 2,000 identities.
UFine3C evaluates retrieval across domains, text granularities, and
writing styles. It is formed by sampling 750 identities from the test
sets of each of CUHK-PEDES, ICFG-PEDES, and UFine6926, representing
coarse-, medium-, and fine-grained descriptions, respectively. Qwen-14B
and Llama2-70B are used to restructure the original descriptions, and
the generated text is manually revised. UFine3C contains 2,250
identities, 7,446 images, and 37,939 text queries.

\subsubsection{Implementation details.}
Following previous works~\citep{Bi-IRRA}, GradCO is initialized with the encoder weights of X2VLM~\citep{X-VLM}. We use the AdamW optimizer with a batch size of $22$ and a weight decay of $0.01$. B-MoE contains $K=4$ sub-experts, and direction-diversity regularization weight is set to $\gamma=0.5$. All experiments are conducted on a NVIDIA V100 GPU with $32$GB memory and trained for $10$ epochs .

\noindent \textbf{Evaluation.} For TPAS, we report Recall@K (R@K, $K\in\{1,5,10\}$) and mean Average Precision (mAP). For TIPR, R-Avg is the arithmetic mean of R@1, R@5, and R@10. On Multi-Weather, we report R@1 and mAP for each of the ten weather conditions and their arithmetic means. PAB-Avg and UCC(OOD)-Avg follow the corresponding protocols. Higher scores indicate better retrieval performance.

\subsection{Performance Comparison}
We compare GradCO with existing methods on the TPAS and TIPR tasks. For TPAS, we evaluate in-distribution retrieval on PAB, robustness under the Multi-Weather protocol, and transfer to the UCC out-of-distribution (OOD) test set. We then evaluate whether the same model remains effective on conventional TIPR benchmarks.

\begin{figure*}[t]
  \centering
\includegraphics[width=0.95\linewidth]{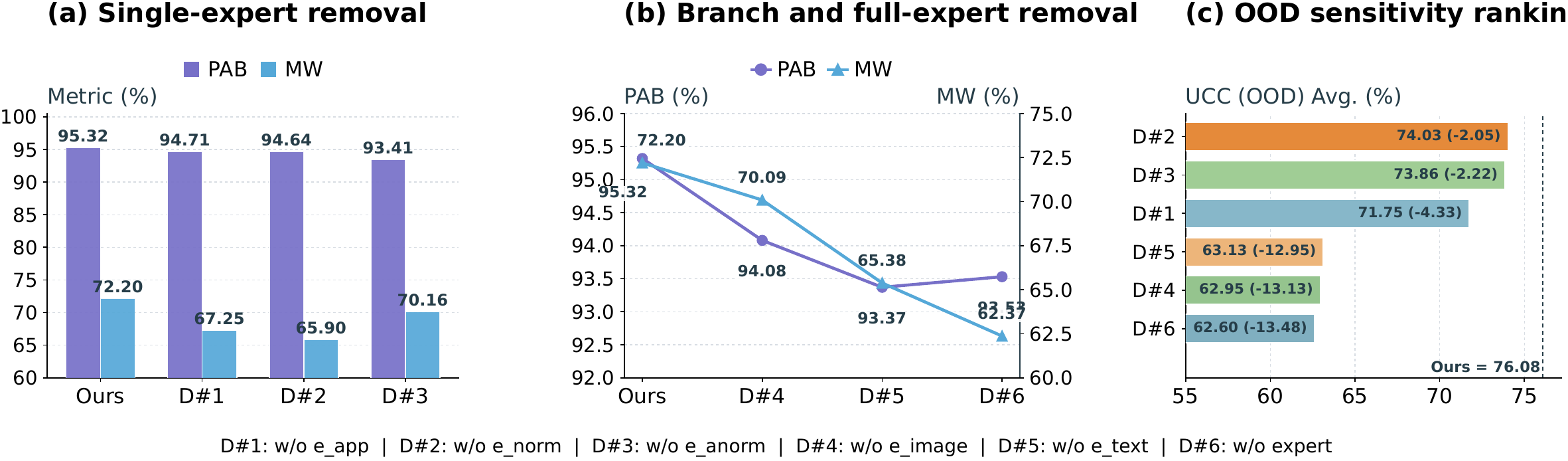}
   \caption{Ablation results for the B-MoE module. (a) Single-expert removal, (b) branch and full-expert removal, and (c) UCC performance ranking across ablation variants. Variant definitions are listed below the panels.}
   \label{fig:GradCO-Ablation-1}
\end{figure*}

\subsubsection{TPAS task}
\noindent \textbf{PAB evaluation.} Table~\ref{tab:pab} reports the results on PAB under the $0.1$M and $1$M gallery settings. GradCO achieves the highest R@1, R@5, and mAP in both settings, and ties CMP at $1$M R@10. Under the $0.1$M setting, GradCO improves R@1 from $83.06$ to $85.39$ and mAP from $90.41$ to $92.13$ over CMP. Under the $1$M setting, the corresponding gains are $1.67$ and $1.22$ percentage points. These results show that GradCO improves text-to-image retrieval when the query combines identity-related appearance with normal or anomalous behavior.

\noindent \textbf{Multi-Weather evaluation.} 
Table~\ref{tab:multiweather} reports R@1 and mAP under ten weather conditions. GradCO achieves the highest value for both metrics in every condition, including the challenging ``Dark+Rain'' setting. Its mean R@1 and mAP are $72.2$ and $80.67$, compared with $67.12$ and $77.44$ for CMP. The consistent gains across conditions indicate that the multistream representation and query-conditioned aggregation remain effective when the visual evidence is degraded by weather changes.

\noindent \textbf{OOD evaluation.} 
Table~\ref{tab:ood} evaluates transfer to UCC. With the $1$M gallery, GradCO improves over CMP from $55.23$ to $62.53$ in R@1 and from $44.35$ to $52.01$ in mAP, corresponding to gains of $7.30$ and $7.66$ percentage points. The improvement is also observed in the $1$M gallery setting, showing that GradCO transfers more effectively to the tested OOD domain.

\subsubsection{TIPR task}
\noindent \textbf{TIPR evaluation.} 
Table~\ref{tab:tipr} summarizes the average R@K on five TIPR evaluation sets. GradCO obtains the highest score on CUHK-PEDES, ICFG-PEDES, RSTPReID, and UFine3C. On UFine6926, Bi-IRRA reaches $95.18$, while GradCO obtains $95.13$. The largest improvement appears on UFine3C, where GradCO reaches $82.01$ compared with $74.63$ for CFAM. These results show that the appearance--behavior modeling and intent-conditioned matching developed for TPAS also improve retrieval under conventional and cross-domain TIPR settings.

\subsection{Efficiency Evaluation Analysis}
\label{sup:quantitative_efficiency}
To assess efficiency and resource consumption, we compare CMP and GradCO
under the same hardware conditions and batch size of $22$. Results are
reported in Table~\ref{tab:efficiency_comparison}.

GradCO has 247.36M parameters, compared with 226.13M for CMP. This
increase mainly comes from B-MoE, which processes appearance, normal
behavior, and anomalous behavior through separate expert streams and
their sub-experts. This design is associated with retrieval scores of
$95.32\%$ on PAB and $76.08\%$ on UCC (OOD).

GradCO uses 18,413 MiB of GPU memory, compared with 14,575 MiB for CMP.
This higher training memory consumption is partly due to COL, which
computes pair-conditioned offload anchors for expert streams during
training to mitigate gradient conflicts caused by ``same identity,
different behavior'' hard negatives.

At inference, GradCO takes 2.3 ms per sample, compared with 1.5 ms for
CMP, an absolute increase of 0.8 ms (53.3\%). This overhead mainly comes
from HIA, which uses text-conditioned gating to aggregate expert
similarities. Despite the increase, the 2.3 ms per-sample retrieval
latency indicates that GradCO can support real-time retrieval in
text-based person anomaly search. Meanwhile, Multi-Weather-Avg improves
from $67.12\%$ to $72.19\%$, and UCC (OOD)-Avg from $68.30\%$ to
$76.08\%$. These results show that GradCO improves retrieval performance
under weather corruptions and distribution shifts while maintaining
low-latency inference, making it suitable for integration into
real-time anomaly-search systems.

\subsection{Ablation Studies}
We evaluate the contributions of B-MoE, COL, and HIA under the common training and evaluation settings described in Sec.~\ref{sec:experimental_settings}. All values are compared with the complete GradCO model.

\begin{figure*}[t!]
  \centering
\includegraphics[width=\linewidth]{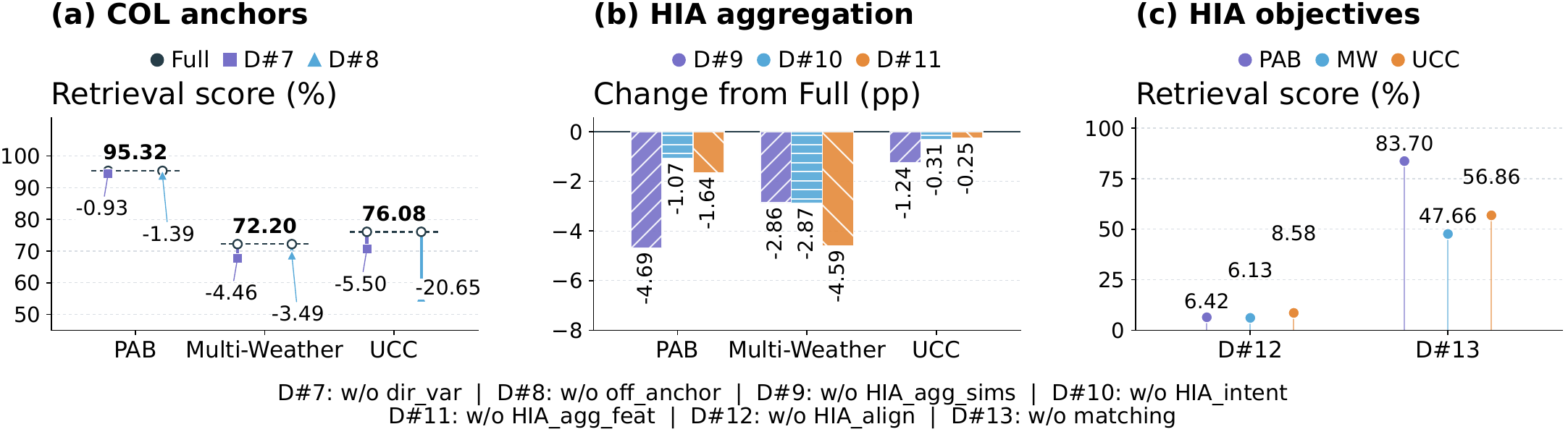}
   \caption{Ablation results for COL and HIA. (a) Effects of removing COL components, with arrows indicating score changes relative to Ours. (b) Performance changes after removing HIA aggregation components. (c) Retrieval performance after removing HIA training objectives. Variant definitions are listed below the panels.}
   \label{fig:GradCO-Ablation-2}
\end{figure*}

\begin{figure}[h]
  \centering
   \resizebox{\linewidth}{!}{\includegraphics[width=\linewidth]{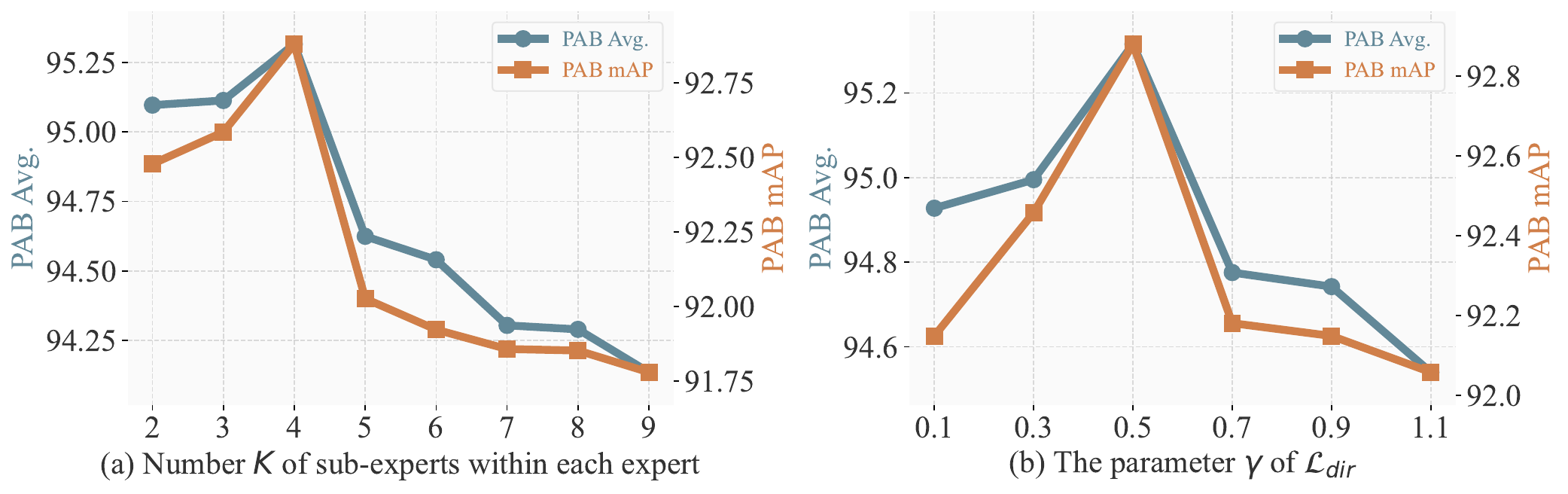}}
   \caption{Sensitivity to the hyper-parameters (a) number of sub-experts $K$, and (b) $\gamma$ on PAB datasets.}
   \label{fig:sentivity}
\end{figure}

\noindent \textbf{B-MoE.}
As shown in Fig.~\ref{fig:GradCO-Ablation-1}, removing the appearance (D\#1), normal-behavior (D\#2), and anomaly-behavior (D\#3) streams reduces PAB-Avg to 94.71, 94.64, and 93.41, respectively, and Multi-Weather-Avg to 67.25, 65.90, and 70.16. Removing the image-side experts (D\#4) or text-side experts (D\#5) lowers UCC-Avg to 62.95 and 63.13; removing B-MoE entirely (D\#6) further reduces it to 62.60. These complementary gains indicate that appearance, normal-behavior, and anomaly-behavior information jointly improve retrieval performance.

\noindent \textbf{COL.}
As shown in Fig.~\ref{fig:GradCO-Ablation-2}(a), removing the directional diversity regularizer (D\#7) lowers UCC-Avg to 70.58, while removing the offload-anchor path (D\#8) further reduces it to 55.43. The larger performance drop indicates that the pair-conditioned anchor is the primary mechanism for mitigating gradient conflicts between samples with the same identity but different behaviors. The directional diversity regularizer helps prevent anchor collapse and maintain effective gradient offloading.

\noindent \textbf{HIA.}
As shown in Fig.~\ref{fig:GradCO-Ablation-2}(b), removing similarity aggregation (D\#9, w/o HIA\_agg\_sims) reduces PAB-Avg by 4.69 points. Removing intent weighting (D\#10, w/o HIA\_intent) and feature aggregation (D\#11, w/o HIA\_agg\_feat) primarily affects PAB and Multi-Weather, indicating that expert information should be dynamically aggregated according to text intent. Removing the image-text contrastive objective (D\#12, w/o HIA\_align) causes performance to nearly collapse. Removing the matching objective (D\#13, w/o matching) reduces PAB-Avg, Multi-Weather-Avg, and UCC-Avg to 83.70, 47.66, and 56.86. These results show that intent aggregation and both training objectives jointly support retrieval performance.

\subsection{Sensitivity Analysis}
To analyze the sensitivity of GradCO to the number of B-MoE sub-experts and the direction-diversity weight $\gamma$, we vary both hyperparameters and report the results in Figure~\ref{fig:sentivity}. For both parameters, the performance first increases and then decreases. Setting $K=4$ provides sufficient perspectives for separating appearance and behavior cues, whereas a larger value (e.g., $K\geq5$) may over-partition features and make experts focus on local details. The best performance is obtained at $\gamma=0.5$, which helps prevent anchor collapse. A larger $\gamma$ over-constrains anchor directions, while a smaller value provides insufficient regularization, leading to less reliable offloading and stronger residual conflicts. We use $K=4$ and $\gamma=0.5$ in all experiments.

\subsection{Mechanism Analysis}

\subsubsection{Performance Analysis on PAB Test Subsets (Normal vs. Anomaly)}
\label{sup:quantitative_normal_anomaly}

\begin{table}[t]
  \centering
  \caption{Performance comparison between GradCO and the baseline model CMP on the two subsets (normal and anomaly) of PAB. GradCO is highlighted in light purple, and the overall best results are shown \textbf{in bold}.}
  \tabcolsep=12pt
  \resizebox{\linewidth}{!}{%
    \begin{tabular}{c|cc|cc}
    \Xhline{1.5pt}
    \multirow{2}{*}{Method} & \multicolumn{2}{c|}{PAB-Normal} & \multicolumn{2}{c}{PAB-Anomaly} \\
\cline{2-5}          & R@1   & mAP   & R@1   & mAP \\
    \hline
    CMP   & 85.04 & 92.04 & 83.22 & 90.24 \\
    \rowcolor[rgb]{0.949, 0.949, 1.0} \textbf{GradCO (Ours)}  &  \textbf{86.55}    & \textbf{92.82}    & \textbf{86.65}  & \textbf{92.94} \\
    \Xhline{1.5pt}
    \end{tabular}%
    }
  \label{tab:pab subset}%
\end{table}%

To validate the effectiveness of GradCO in addressing the ``invariance'' versus ``sensitivity'' conflict inherent in feature learning, we partitioned the PAB test set into two subsets: PAB-Normal (regular behaviors) and PAB-Anomaly (anomalous behaviors). This division aims to separately evaluate the decoupling capability of the model in terms of identity preservation and fine-grained action capture. As shown in Table \ref{tab:pab subset}, compared to the baseline model CMP, GradCO achieved performance improvements across all metrics on both subsets, with the specific analysis as follows:

\begin{figure}[h]
  \centering
    \resizebox{\linewidth}{!}{\includegraphics[width=\linewidth]{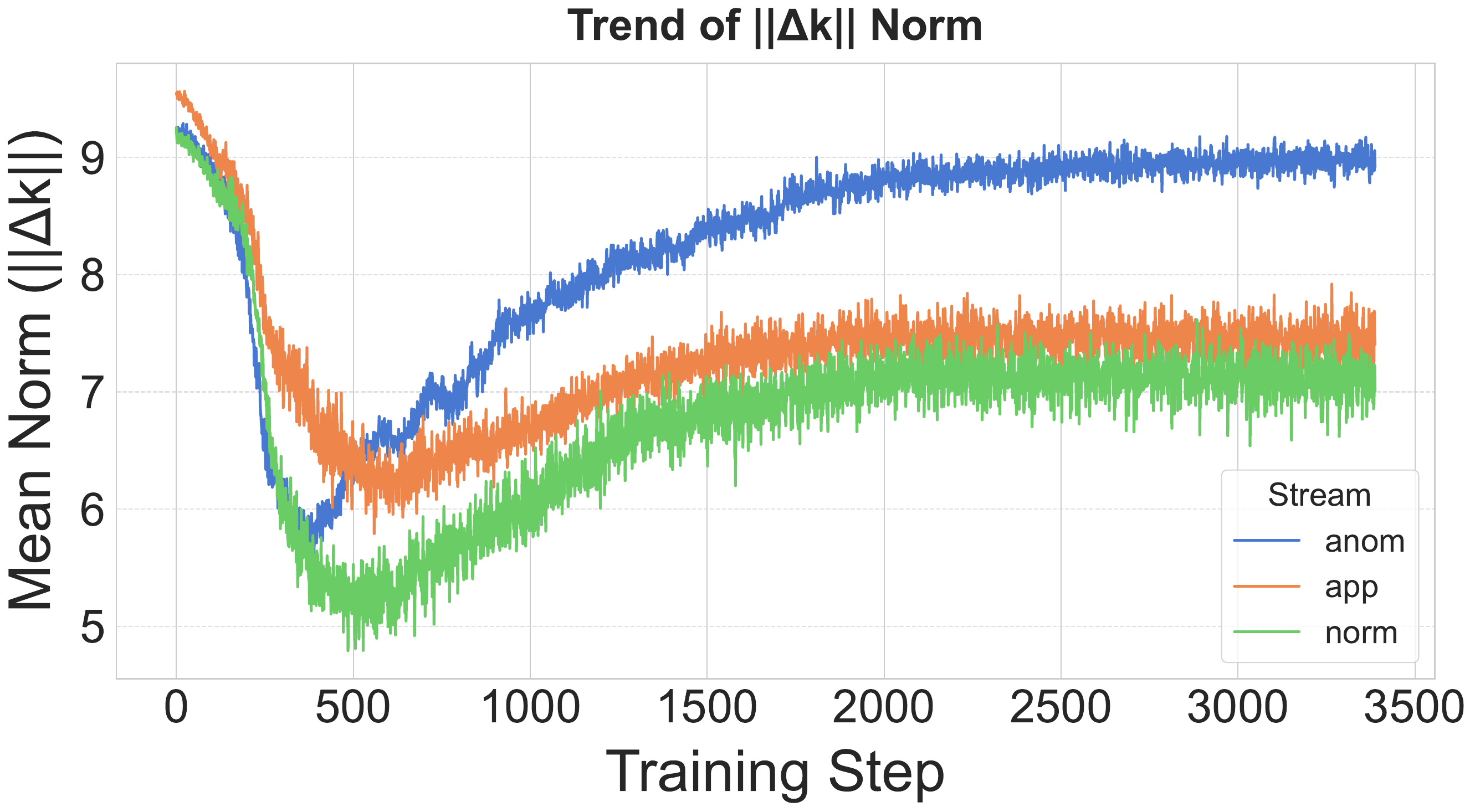}}
   \caption{Evolution of the offload anchor norm ($||\Delta_k||$) in the COL module during training. The curves exhibit a trend of initial decline followed by a rise and stabilization. Notably, the anomaly behavior stream (anom) consistently maintains a higher magnitude compared to the appearance (app) and normal behavior (norm) streams.
   }
   \vspace{-15pt}
   \label{fig:trend of deltak norm step}
\end{figure}

\textbf{1)} On the PAB-Normal subset, GradCO elevates the R@1 metric from the baseline of $85.04\%$ to $86.55\%$. This demonstrates that after the Behavior-Aware Mixture-of-Experts (B-MoE) disentangles highly coupled semantics into independent expert streams, it does not compromise the capability of the model to extract appearance features. Coupled with Conflict Offload Learning (COL) for the mitigation of gradient conflict among experts, the model maintains robust identity recognition performance when processing regular behaviors.

\textbf{2)} On the PAB-Anomaly subset, the performance of CMP exhibits a decline, with its R@1 being only $83.22\%$ (lower than its performance on PAB-Normal). This is attributed to its traditional ``Aggregate-then-Align'' paradigm causing ``Aggregation Distortion'', which renders subtle anomaly behavior signals susceptible to being overwhelmed by strong appearance features. In contrast, GradCO achieves an R@1 of $86.65\%$ and an mAP of $92.94\%$ on this subset. Benefiting from the intent-guided dynamic aggregation of Hierarchical Intent Alignment (HIA) and the offload anchors mechanism of COL, the model successfully preserves and precisely aligns crucial evidence of anomaly behavior without the need for additional supervision signals.

\subsubsection{The Evolution Analysis of Offload Anchors $\Delta$}
\label{sup:quantitative_Delta}

To characterize the evolution of COL's offload anchors during training,
we track their norms and their cosine similarity with the text-image
semantic difference across the appearance, normal-behavior, and
anomalous-behavior streams. These metrics describe the anchors' magnitude
and direction relative to the main alignment signal.

\begin{figure}[ht]
  \centering
    \resizebox{\linewidth}{!}{\includegraphics[width=\linewidth]{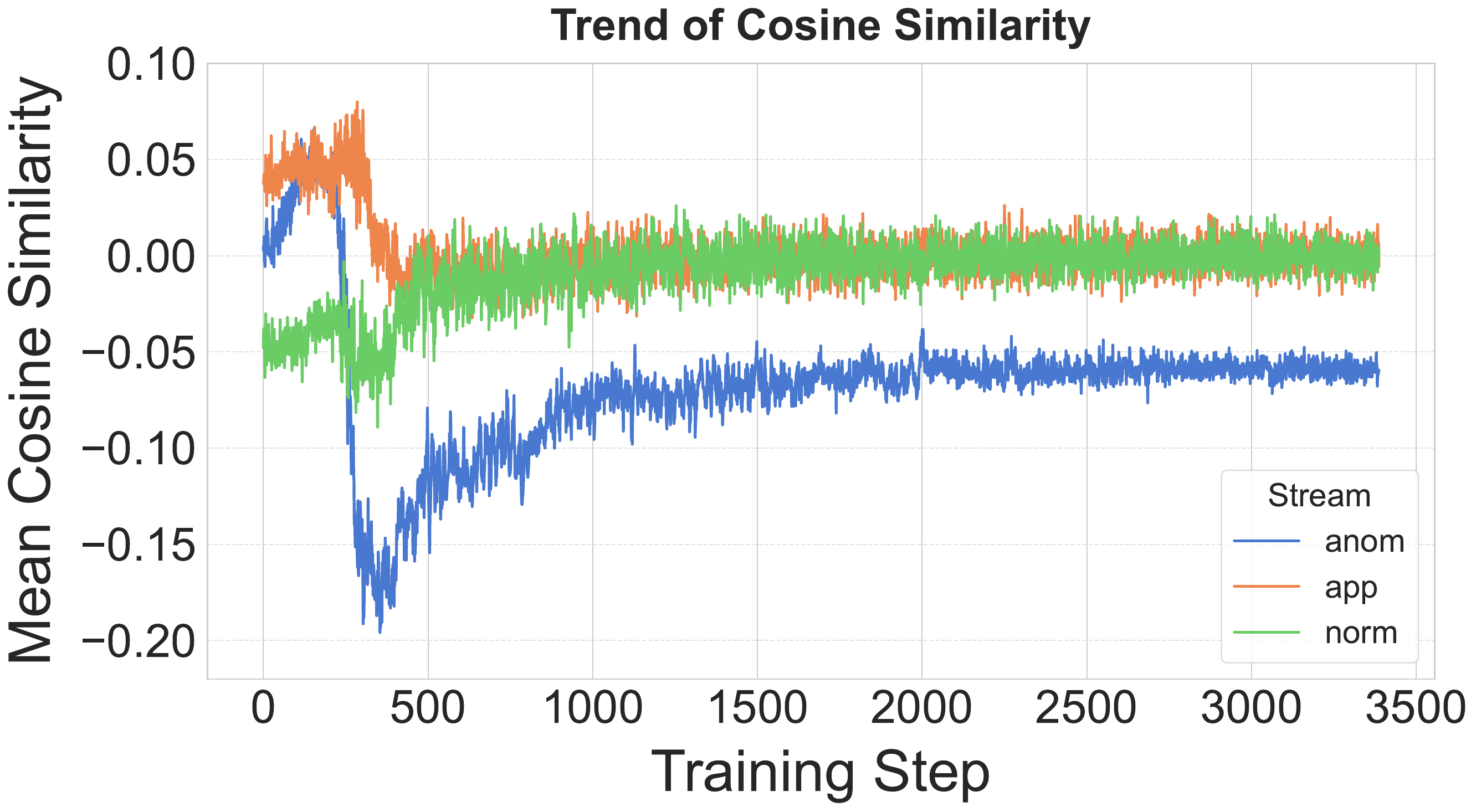}}
   \caption{Trend of cosine similarity between offload anchors $\Delta_k$ and the main feature semantic difference. The similarities for all streams stabilize near 0, indicating that $\Delta_k$ remains approximately orthogonal to the main feature subspace. The anomaly stream (anom) shows a consistent, slight negative deviation (approximately -0.1) compared to the other streams.
   }
   \label{fig:trend of similarity}
\end{figure}

\begin{figure*}[t]
    \centering
    \includegraphics[width=\textwidth]{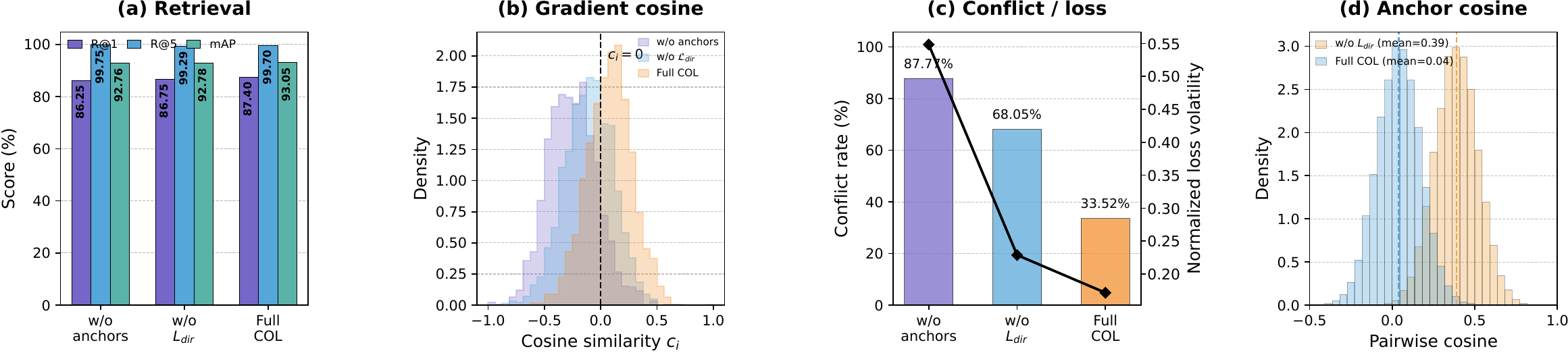}
    \caption{
    Mechanistic analysis of how COL resolves identity--behavior conflicts.
    (a) Retrieval performance comparison among different COL variants.
    (b) Gradient cosine similarity between identity-preserving and
    behavior-discriminative optimization signals.
    (c) Optimization stability measured by normalized loss volatility.
    (d) Directional diversity of learned offload anchors.
    }
    \label{fig:col_mechanism}
\end{figure*}

Figure~\ref{fig:trend of deltak norm step} shows that anchor norms first
decrease, then rise, and finally stabilize. During the first 500 steps,
the model builds a coarse global feature space, and the anchors remain
small. From steps 500 to 2,000, as training focuses more on hard
negatives with the same identity but different behaviors, anchor norms
increase, consistent with COL adapting its correction strength to
fine-grained alignment conflicts. The anomaly stream maintains the
largest norm (approximately 9.0), followed by the appearance stream
(approximately 7.5) and the normal-behavior stream (approximately
7.0). This pattern is consistent with the greater difficulty of
preserving subtle anomaly cues against strong appearance signals.

Figure~\ref{fig:trend of similarity} shows that the cosine similarity
between the anchors and the text-image semantic difference remains near
zero throughout training, indicating that the anchors tend to introduce
corrections in directions distinct from the main alignment signal. The
anomaly stream has a slight negative offset (approximately $-0.1$),
suggesting a weak opposing component for this stream. Together, these
trends show that COL learns stable, stream-dependent corrections while
preserving the main semantic alignment direction. The stable late-stage
curves are also consistent with $\mathcal{L}_{dir}$ helping prevent
anchor-direction collapse.

\begin{table}[h]
\centering
\caption{
Optimization diagnostics of COL variants.
Gradient conflict rate and loss volatility measure optimization interference,
while retrieval margin and anchor directional similarity characterize
retrieval separation and offload geometry, respectively.
}
\label{tab:col_diagnostics}
\begin{tabular}{c|cccc}
\hline
Method &
$r_{\mathrm{conf}}\downarrow$ &
$V_{\mathcal L}\downarrow$ &
$m\uparrow$ &
$\mu_{\Delta}\downarrow$\\
\hline
w/o anchors &
87.0\% &
0.072 &
0.18 &
--\\

w/o $\mathcal L_{dir}$ &
45.4\% &
0.043 &
0.46 &
0.681\\

Full COL &
12.2\% &
0.019 &
0.78 &
0.129\\
\hline
\end{tabular}
\end{table}

\begin{figure}[t]
  \centering
   \resizebox{\linewidth}{!}{\includegraphics[width=\linewidth]{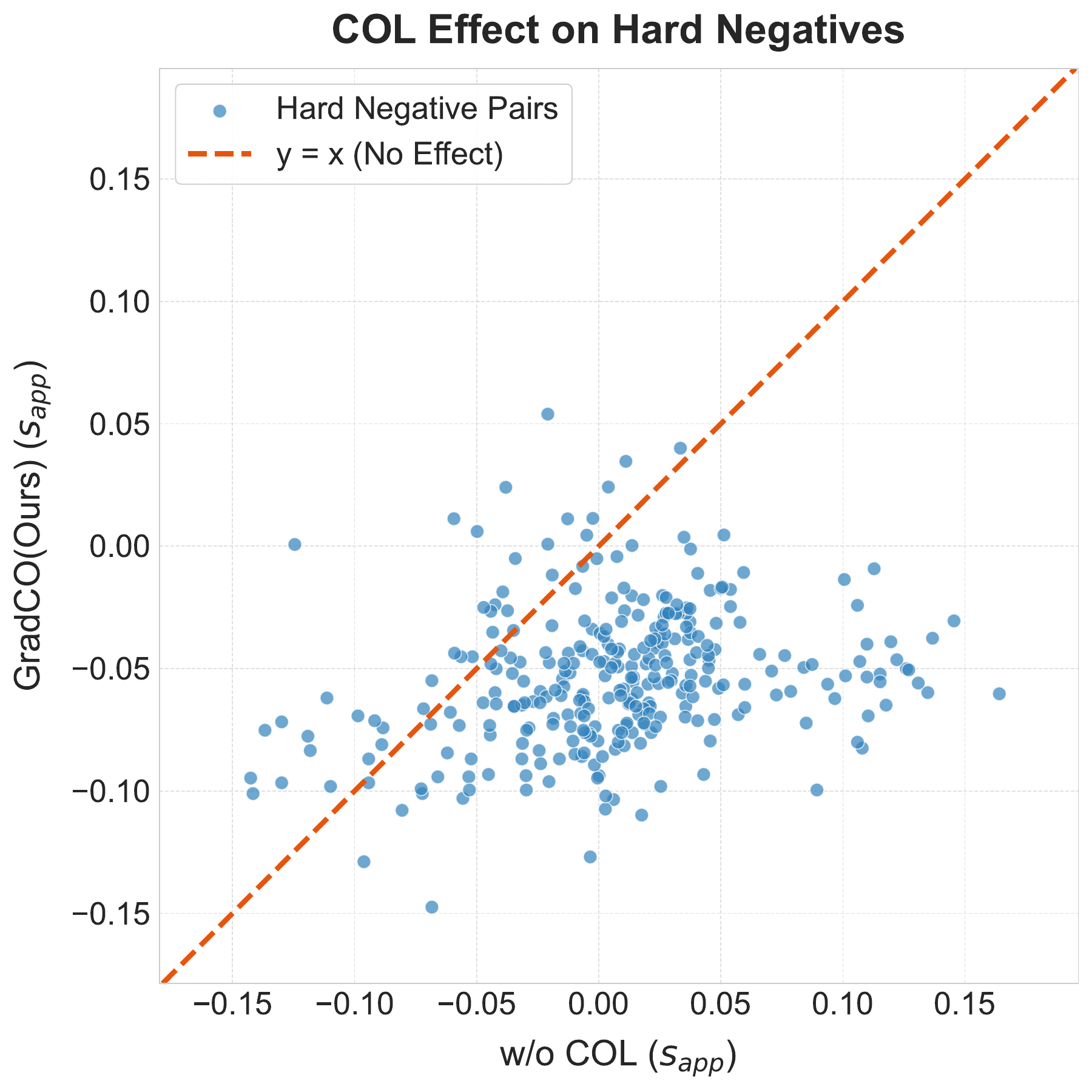}}
   \caption{
Instance-level correction of hard negatives by COL.
The scatter plot compares appearance-expert similarity scores before
and after applying COL on the same hard negative pairs.
Most samples fall below the diagonal $y=x$, indicating that COL suppresses
confusing similarity responses.
}
   \label{fig:similarity_scatter}
\end{figure}

\begin{figure*}[t]
  \centering
  \includegraphics[width=0.99\textwidth]{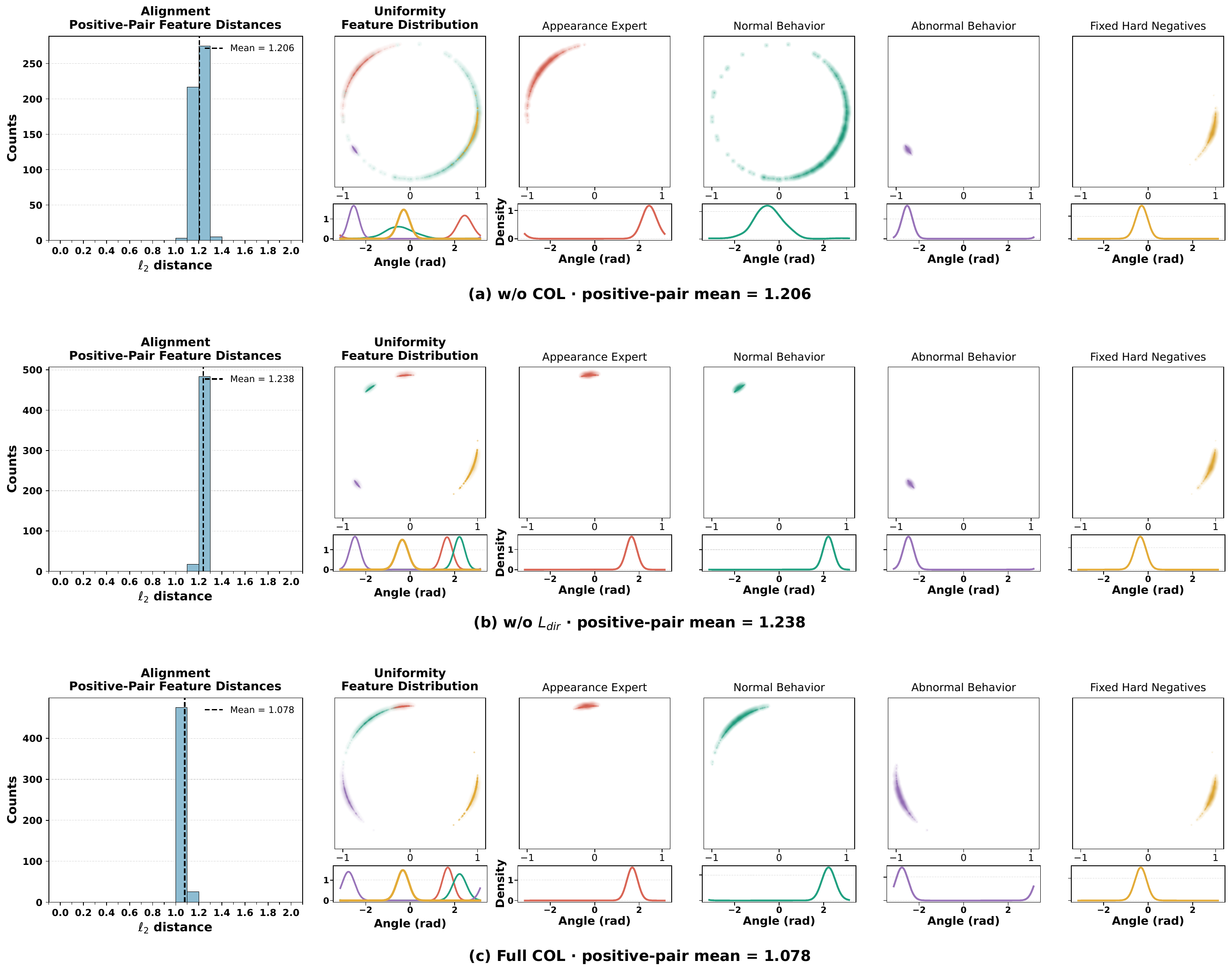}
  \caption{
  Feature alignment and embedding geometry analysis of COL.
  Each row corresponds to (a) w/o COL, (b) COL w/o $\mathcal{L}_{dir}$,
  and (c) Full COL. The left column shows positive-pair text-image feature
  distances. The middle column presents the global feature distribution and
  angular density. The right columns show the projected distributions of
  Appearance Expert, Normal Behavior, Abnormal Behavior, and fixed hard
  negatives.
  }
  \label{fig:col_alignment_uniformity_kde}
\end{figure*}

\subsubsection{How COL Resolves Identity--Behavior Conflicts}
\label{sup:ablation_col_effectiveness}
This section analyzes the optimization mechanism of COL from three perspectives:
gradient interaction, optimization stability, and anchor geometry. In TPAS,
hard negatives often share similar identity-related appearance information
while exhibiting different behaviors, creating competing optimization signals
between identity preservation and behavior discrimination.

\begin{figure*}[t]
  \centering
  \includegraphics[width=0.99\textwidth]{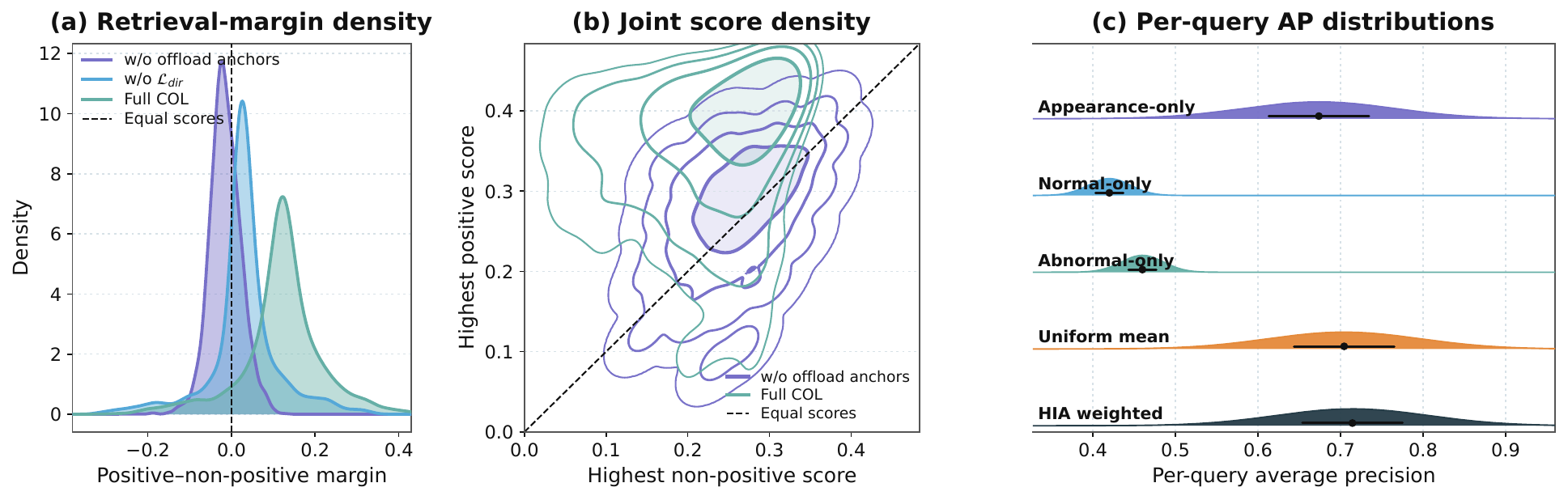}
  \caption{
  Retrieval separation and adaptive expert aggregation analysis.
  (a) KDE of query-level retrieval margins
  $m_i=s_i^+-s_i^-$ under different COL variants.
  (b) Joint density of the highest positive and non-positive scores,
  illustrating the score redistribution induced by COL.
  (c) Query-level AP distributions of individual experts, uniform averaging,
  and HIA-weighted aggregation.
  }
  \label{fig:retrieval_margin_expert_aggregation}
\end{figure*}

COL introduces sample-specific offload anchors to provide an additional
optimization pathway for conflict-sensitive information. For expert branch $k$,
the anchor is generated as
\begin{equation}
\Delta_{ij}^{k}
=
TF_k(Q=f^{T}_{k,i}-f^{I}_{k,j},
K,V=f^{I}_{k,j}),
\end{equation}
and the similarity is reformulated as
\begin{equation}
S_{ij}^{k}
=
\cos(f^{T}_{k,i}+\Delta_{ij}^{k},f^{I}_{k,j}),
\end{equation}
where $\Delta_{ij}^{k}$ provides a sample-specific correction direction while
preserving the original feature pathway for identity-related representation.

\noindent \textbf{Gradient conflict reduction.}
We quantify the interaction between positive attraction and hard-negative
repulsion by measuring their gradient cosine similarity. Let
$\mathbf g_i^+$ and $\mathbf g_i^-$ denote the corresponding gradients
propagated to the same text-side representation:
\begin{equation}
c_i=
\frac{
\langle \mathbf g_i^+,\mathbf g_i^-\rangle
}
{
\|\mathbf g_i^+\|_2\|\mathbf g_i^-\|_2+\epsilon
}.
\end{equation}
A negative cosine value indicates conflicting optimization directions.
The conflict rate is computed as
\begin{equation}
r_{\mathrm{conf}}
=
\frac1N\sum_i\mathbb{1}[c_i<0].
\end{equation}
The hardest negative sample is selected from non-matching candidates within
the same mini-batch according to the highest similarity score.

\noindent \textbf{Optimization stability and retrieval separation.}
To evaluate optimization stability, we measure normalized loss volatility:
\begin{equation}
V_{\mathcal L}
=
\frac{
\operatorname{Std}(\{\mathcal L_t\}_{t\in\mathcal W})
}
{
\operatorname{Mean}(\{\mathcal L_t\}_{t\in\mathcal W})+\epsilon
}.
\end{equation}
Meanwhile, retrieval separation is quantified by the gallery-level margin:
\begin{equation}
m_i=
\max_{j\in\mathcal P_i}S_{ij}
-
\max_{j\notin\mathcal P_i}S_{ij}.
\end{equation}

\noindent \textbf{Anchor diversity preservation.}
To investigate the effect of directional regularization, the normalized anchor
direction is defined as
\begin{equation}
d_{ij}
=
\frac{\Delta_{ij}}
{\|\Delta_{ij}\|_2+\epsilon}.
\end{equation}
The average directional similarity is calculated as
\begin{equation}
\mu_{\Delta}
=
\frac1{|\mathcal Q|}
\sum_i
\frac1{|\mathcal P_i^\Delta|}
\sum_{(j,k)}
\langle d_{ij},d_{ik}\rangle .
\end{equation}
Lower $\mu_\Delta$ indicates more diverse anchor directions.

As shown in Fig.~\ref{fig:col_mechanism} and
Table~\ref{tab:col_diagnostics}, COL reduces gradient conflicts, stabilizes
optimization, and improves retrieval separation. Removing anchors results in
stronger optimization interference, while removing $\mathcal L_{dir}$ leads
to less diverse anchor directions. These results demonstrate that COL
resolves identity--behavior conflicts by introducing adaptive sample-specific
offloading pathways.

\noindent \textbf{Instance-level correction of hard negatives.}
Beyond optimization-level analysis, we further examine the instance-level
effect of COL on confusing hard negative samples. Figure~
\ref{fig:similarity_scatter} compares the appearance-expert similarity scores
of the same hard negative pairs before and after introducing COL.
Most samples are distributed below the diagonal line $y=x$, indicating that
COL consistently reduces the similarity assigned to hard negatives. This
sample-level correction reduces the tendency of behavior-conflicting samples
to compete with true matches in the retrieval space, providing additional
evidence that the offload anchors translate optimization benefits into
improved discrimination of difficult cases.

\subsubsection{How COL Improves Feature Alignment and Embedding Geometry}
\label{sup:col_alignment_uniformity}
Beyond optimization dynamics, we further investigate how COL affects the
geometry of the learned embedding space. Specifically, we analyze three
aspects: cross-modal positive-pair alignment, global feature distribution, and
the distribution structure of different semantic groups. 
Figure~\ref{fig:col_alignment_uniformity_kde} compares the embedding geometry
under three settings: w/o COL, COL w/o $\mathcal{L}_{dir}$, and Full COL.
All variants use identical feature extraction, projection, and density
estimation procedures.

\begin{figure*}[t]
  \centering
  \includegraphics[width=0.99\textwidth]{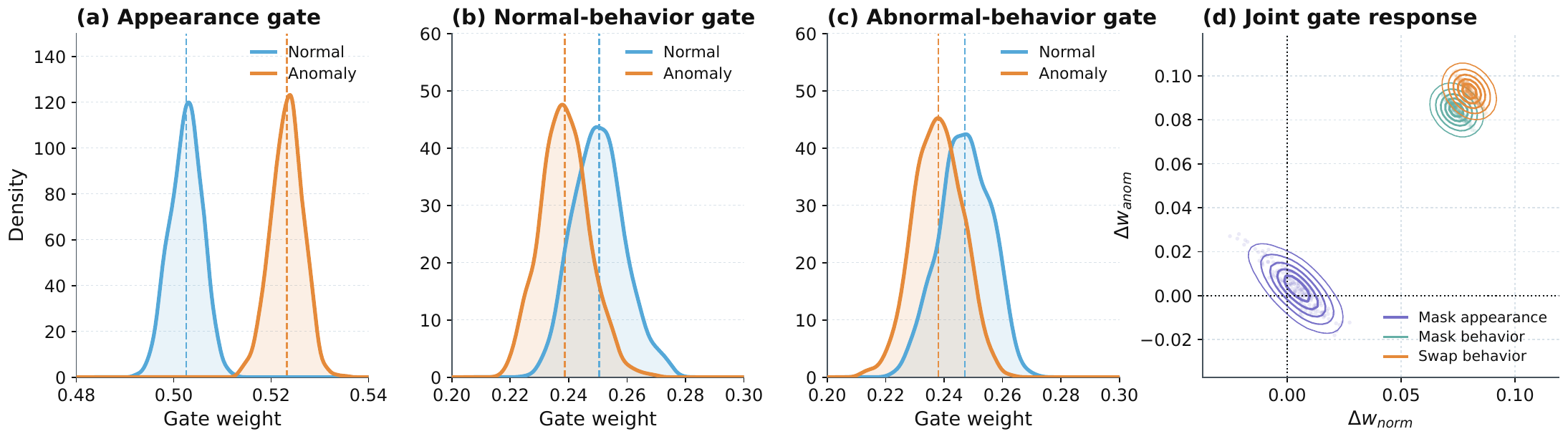}
  \caption{
  Query-conditioned expert activation analysis of HIA.
  (a--c) KDE distributions of appearance, normal-behavior, and
  abnormal-behavior gate weights for official Normal and Anomaly queries.
  Dashed lines indicate group means.
  (d) Joint density of paired gate variations under controlled text
  interventions, including appearance masking, behavior masking, and
  behavior replacement.
  }
  \label{fig:hia_activation_text_intervention}
\end{figure*}

For each ground-truth text-image pair, let
$\mathbf z_i^{\mathbb T}$ and $\mathbf z_i^{\mathbb I+}$ denote the
$\ell_2$-normalized text and image embeddings, respectively. The positive-pair
alignment distance is defined as
\begin{equation}
d_i^{+}
=
\left\|
\mathbf z_i^{\mathbb T}
-
\mathbf z_i^{\mathbb I+}
\right\|_2,
\qquad
\bar d_{+}
=
\frac1{N_+}
\sum_{i=1}^{N_+}d_i^{+}.
\label{eq:col_positive_alignment_distance}
\end{equation}
A smaller $\bar d_{+}$ indicates better cross-modal alignment between matched
text and image representations.

To characterize the global embedding geometry, we further measure feature
uniformity in the original normalized feature space:
\begin{equation}
\mathcal U(\mathbf Z)
=
\log
\left[
\frac{1}{N(N-1)}
\sum_{i\neq j}
\exp
(-2\|\mathbf z_i-\mathbf z_j\|_2^2)
\right].
\label{eq:col_uniformity_metric}
\end{equation}
Following previous analysis, a lower value indicates a more uniformly
distributed embedding space with reduced local concentration.

For visualization, a shared projection matrix $P$ is applied to obtain
two-dimensional representations:
\begin{equation}
\mathbf p_i=P\mathbf z_i,
\qquad
\mathbf q_i=
\frac{\mathbf p_i}
{\|\mathbf p_i\|_2+\epsilon},
\end{equation}
where the angular distribution is estimated using kernel density estimation.
The same projection and KDE configurations are applied to all variants for a
fair comparison.

As shown in Fig.~\ref{fig:col_alignment_uniformity_kde}, Full COL achieves the
smallest positive-pair distance, with the mean distance reduced to $1.078$
compared with $1.206$ for w/o COL and $1.238$ for COL w/o $\mathcal{L}_{dir}$.
This indicates that COL improves the alignment between matched text-image
representations.

Meanwhile, removing $\mathcal{L}_{dir}$ leads to a more concentrated embedding
distribution, suggesting that unconstrained correction directions may reduce
the diversity of learned offsets and distort the feature geometry. In contrast,
Full COL maintains a more balanced angular coverage while preserving the
separation among different semantic groups, including appearance, normal
behavior, abnormal behavior, and hard negatives.

\subsubsection{How Does HIA Improve Query-level Retrieval Consistency?}
\label{sup:retrieval_margin_expert_aggregation}
After analyzing the optimization mechanism and feature geometry of COL,
we further investigate how HIA improves query-level retrieval consistency
through adaptive expert aggregation. All comparisons use the same official
query set, complete gallery, ground-truth annotations, and evaluation
protocol.

To understand the role of expert aggregation, we first analyze the retrieval
behavior of different semantic branches. The three expert branches capture
complementary information, including appearance identity, normal behavior
patterns, and abnormal behavior cues. Since different queries may rely on
different semantic factors, fixed expert weighting may not provide optimal
retrieval performance.

For query $q_i$, let
$S_{ij}^{\mathrm{app}}$,
$S_{ij}^{\mathrm{norm}}$, and
$S_{ij}^{\mathrm{anom}}$
denote the similarity scores produced by the three expert branches. Uniform
aggregation and HIA aggregation are defined as
\begin{equation}
S_{ij}^{\mathrm{uni}}
=
\frac{1}{3}
\sum_{k}
S_{ij}^{k},
S_{ij}^{\mathrm{HIA}}
=
\sum_k w_{ik}S_{ij}^{k},
\label{eq:merged_hia_scores}
\end{equation}
where $k\in\{\mathrm{app,norm,anom}\}$, $w_{ik}$ denotes the query-conditioned HIA weight.

For each query, we calculate the average precision (AP) under individual
experts, uniform aggregation, and HIA aggregation. Figure~
\ref{fig:retrieval_margin_expert_aggregation}(c) presents the corresponding
query-level AP distributions. Individual experts show different strengths
across queries, while uniform aggregation provides a fixed combination of
semantic cues. HIA further improves retrieval consistency by dynamically
adjusting expert contributions according to query characteristics.

We further analyze the resulting ranking behavior through positive-negative
score separation. For query $q_i$, let $\mathcal P_i$ denote the set of
ground-truth positive samples. The highest positive and non-positive scores
are defined as
\begin{equation}
s_i^+
=
\max_{j\in\mathcal P_i}S_{ij},
\qquad
s_i^-
=
\max_{j\notin\mathcal P_i}S_{ij}.
\label{eq:col_joint_scores}
\end{equation}

The retrieval margin is calculated as
\begin{equation}
m_i=s_i^+-s_i^- .
\label{eq:merged_retrieval_margin}
\end{equation}
A larger margin indicates a clearer ranking boundary between correct matches
and confusing negatives. Figure~
\ref{fig:retrieval_margin_expert_aggregation}(a) compares the margin
distribution under different COL settings, while Fig.~
\ref{fig:retrieval_margin_expert_aggregation}(b) visualizes the joint
distribution of $(s_i^-,s_i^+)$. These results characterize how the learned
representations translate into final retrieval decisions.

Overall, the analysis demonstrates that HIA improves query-level retrieval
consistency by adaptively combining complementary expert representations.
Instead of relying on a fixed expert contribution, HIA dynamically adjusts
semantic importance according to each query, resulting in more reliable
retrieval behavior.

\begin{figure*}[t]
  \centering
  \resizebox{\linewidth}{!}{\includegraphics[width=\linewidth]{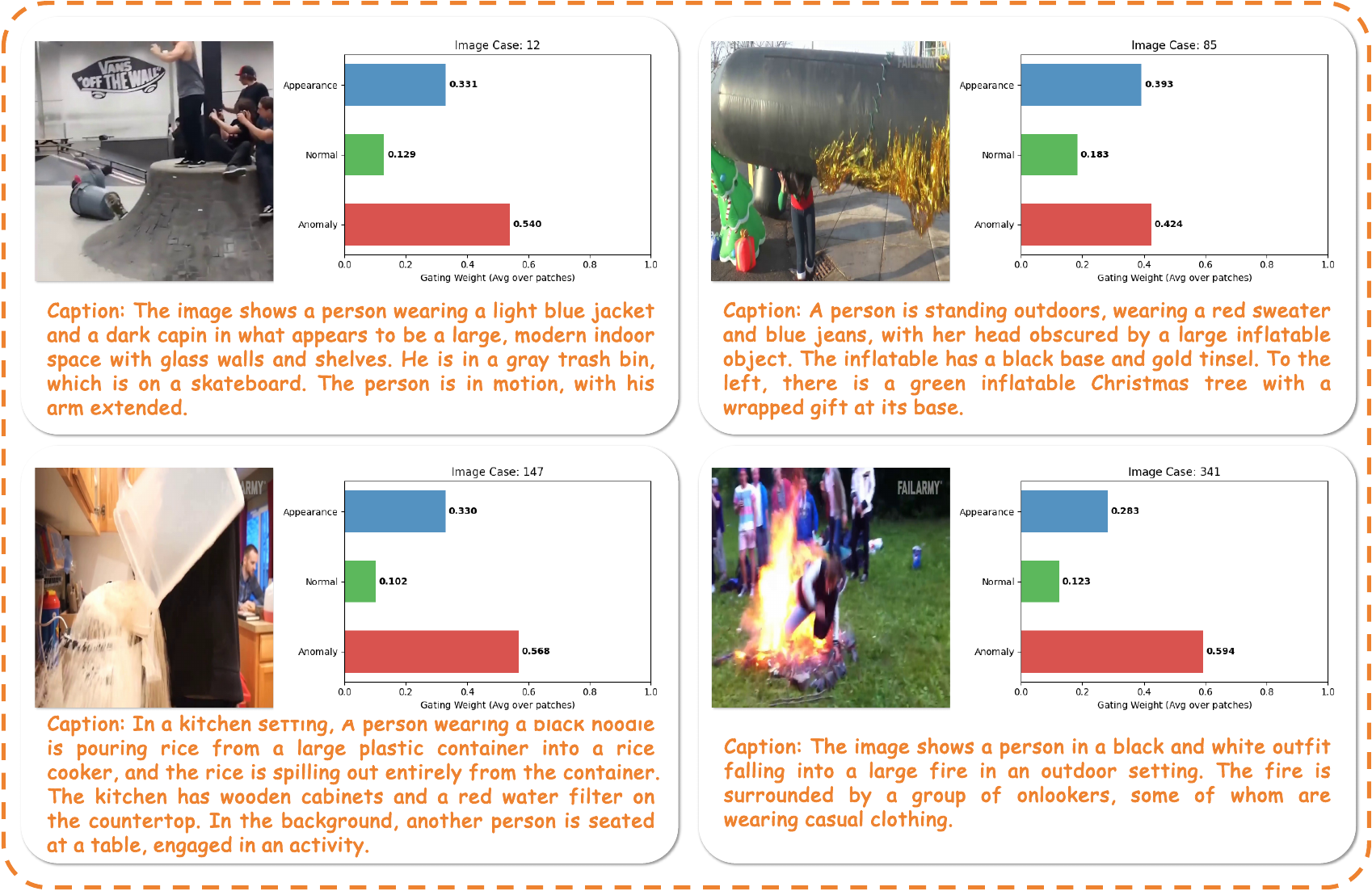}}
  \caption{Representative visualization of B-MoE gating weights for selected anomaly examples. 
  The anomaly-behavior stream receives relatively higher weights in the shown cases, while the appearance stream remains actively weighted. 
  These examples illustrate input-dependent routing preferences learned by the gating network rather than category-level expert assignment.}
  \label{fig:anomaly_case}
\end{figure*}

\begin{figure*}[t]
  \centering
  \resizebox{\linewidth}{!}{\includegraphics[width=\linewidth]{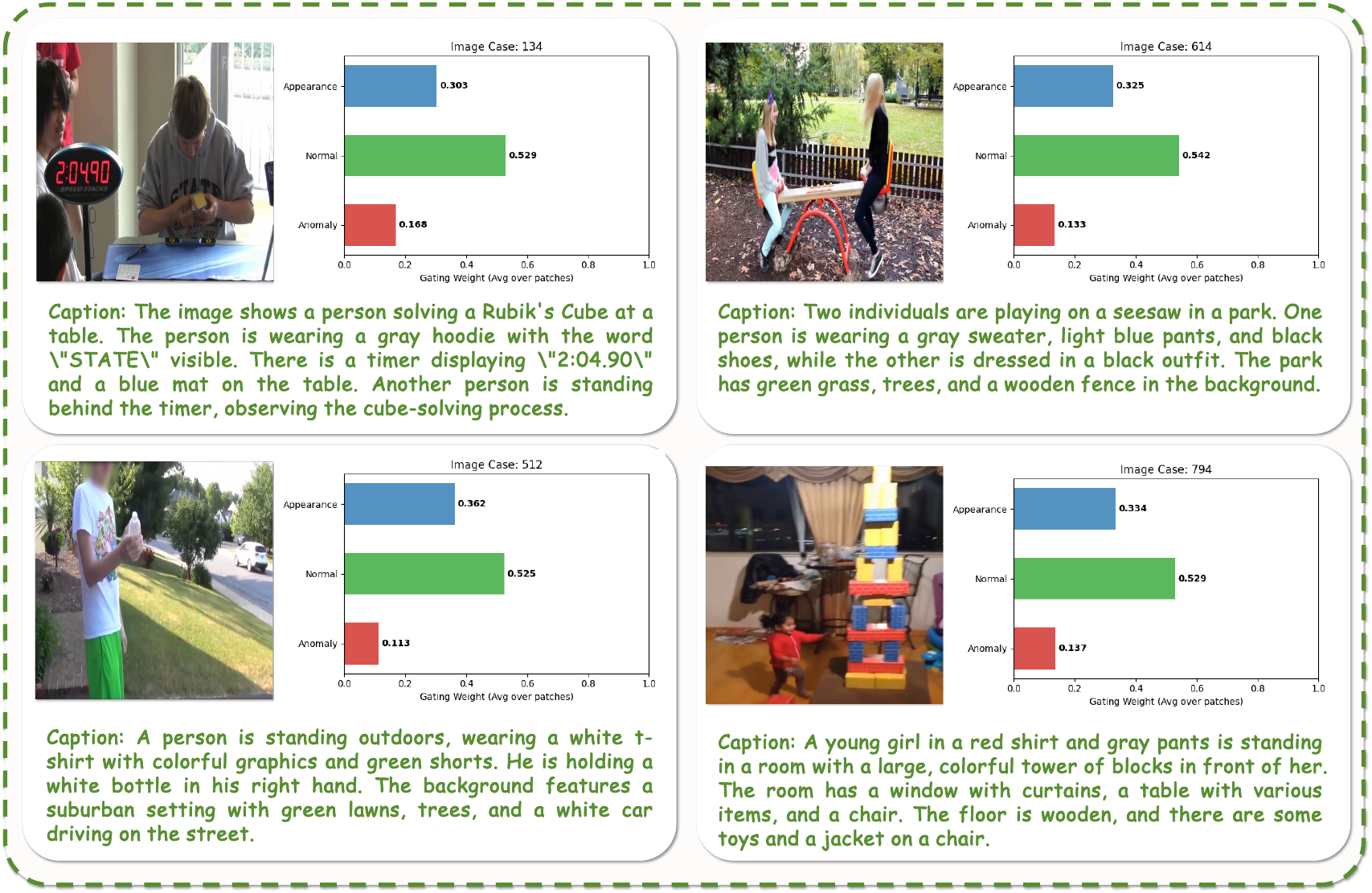}}
  \caption{Representative visualization of B-MoE gating weights for selected normal examples. 
  The normal-behavior stream receives relatively higher weights in the shown controlled or skillful activities, while the anomaly-behavior stream receives lower preferences. 
  The visualization reflects the adaptive routing behavior of B-MoE without implying a predefined semantic boundary among experts.}
  \label{fig:normal_case}
\end{figure*}

\subsubsection{How Does HIA Adapt Expert Weights to Query Semantics?}
\label{sup:hia_activation_text_intervention}
This section analyzes whether HIA performs query-dependent expert adaptation
by examining gate distributions and controlled text interventions. For each
test query, we extract the three expert weights
$\mathbf w_i=(w_{i,\mathrm{app}},w_{i,\mathrm{norm}},w_{i,\mathrm{anom}})$
and analyze their responses under different semantic conditions.

The three weights satisfy
\begin{equation}
w_{i,\mathrm{app}}
+
w_{i,\mathrm{norm}}
+
w_{i,\mathrm{anom}}
=
1,
\qquad
w_{i,k}\in[0,1].
\label{eq:hia_gate_simplex}
\end{equation}

Figure~\ref{fig:hia_activation_text_intervention}(a--c) shows the KDE
distributions of the three expert weights for official Normal and Anomaly
queries. The distributions reveal different routing preferences between the
two query groups: Normal queries assign relatively larger weights to
normal-behavior experts, while Anomaly queries increase the contribution of
abnormal-behavior experts. Meanwhile, the appearance expert maintains a
distinct weighting pattern across the two groups, indicating that HIA
considers both identity and behavior-related semantics during aggregation.

To further examine whether the gate responds to explicit semantic changes,
we perform controlled text interventions. For each query with deterministic
semantic phrases, we construct modified descriptions by masking appearance
phrases, masking behavior phrases, or replacing behavior descriptions. All
interventions preserve the same gallery and evaluation protocol.

For intervention condition $c$, the paired gate variation is defined as
\begin{equation}
\Delta w_{i,k}^{(c)}
=
w_{i,k}^{(c)}
-
w_{i,k}^{(\mathrm{full})}.
\label{eq:hia_gate_intervention_delta}
\end{equation}

Figure~\ref{fig:hia_activation_text_intervention}(d) presents the joint
density of
$\Delta w_{i,\mathrm{norm}}$ and
$\Delta w_{i,\mathrm{anom}}$.
The intervention results show that changing behavior-related descriptions
induces coordinated shifts between normal- and abnormal-behavior expert
weights, whereas appearance masking mainly affects the corresponding routing
response with limited behavior-gate variation. These observations indicate
that HIA dynamically adjusts expert importance according to query semantics
rather than relying on fixed expert aggregation.

Together, these analyses demonstrate that HIA provides adaptive expert
routing by responding to query semantics. The learned gate weights reflect
the relative importance of appearance and behavior cues, enabling flexible
aggregation of complementary expert representations.

\subsection{Qualitative Results}
\label{sup:more_qualitative_results}

\subsubsection{Visualization of B-MoE Expert Gating Weights}
\label{sup:qualitative_moe_gating}

To analyze whether B-MoE learns input-dependent routing patterns, we visualize the
top-level gating weights $\mathbf{w}_{top}^{\mathbb I}$ generated by
$\mathcal{G}_{top}^{\mathbb I}$ in
Figures~\ref{fig:anomaly_case} and~\ref{fig:normal_case}.
The weights correspond to the relative preferences assigned to the
appearance ($\mathbf{U}_{app}^{\mathbb I}$),
normal-behavior ($\mathbf{U}_{norm}^{\mathbb I}$),
and anomaly-behavior ($\mathbf{U}_{anom}^{\mathbb I}$) streams.
Since all experts receive the same encoder input and no expert-specific labels
are provided during training, these visualizations should be interpreted as
learned routing preferences rather than explicit semantic categorization.

In the selected anomaly examples, the anomaly-behavior stream receives
relatively larger weights compared with the other streams.
For example, Case 341 assigns a weight of 0.594 to the anomaly stream,
indicating that the gate increases its preference for this stream under the
shown abnormal visual context.
Meanwhile, the appearance stream remains nonzero, suggesting that the routing
process preserves contributions from multiple expert streams rather than
selecting a single exclusive pathway.

For the selected normal examples, the normal-behavior stream obtains higher
relative weights in scenarios involving controlled or skillful activities.
For instance, in the Rubik's cube example, the normal-behavior stream receives
a weight of 0.529, while the anomaly-behavior stream receives 0.137.
This indicates that B-MoE assigns different routing preferences according to
the observed visual content, allowing different expert streams to contribute
differently across samples.

The appearance stream remains active in the displayed examples, including a
weight of 0.393 in Case 85, while behavior-related streams become more
prominent when stronger action cues are present.
This observation suggests that the learned routing mechanism maintains a
balance among different expert streams instead of relying exclusively on
appearance or behavior information.
Overall, these visualizations provide qualitative evidence that the
B-MoE gating network learns sample-dependent expert preferences.
Together with the quantitative ablation results, they support the motivation
of decomposing TPAS representations into multiple specialized streams, illustrating relative expert preferences rather than
explicit semantic labels assigned to each expert.

\begin{figure*}[ht] 
    \centering
    \begin{subfigure}[b]{0.48\textwidth}
        \centering
        \includegraphics[width=\linewidth]{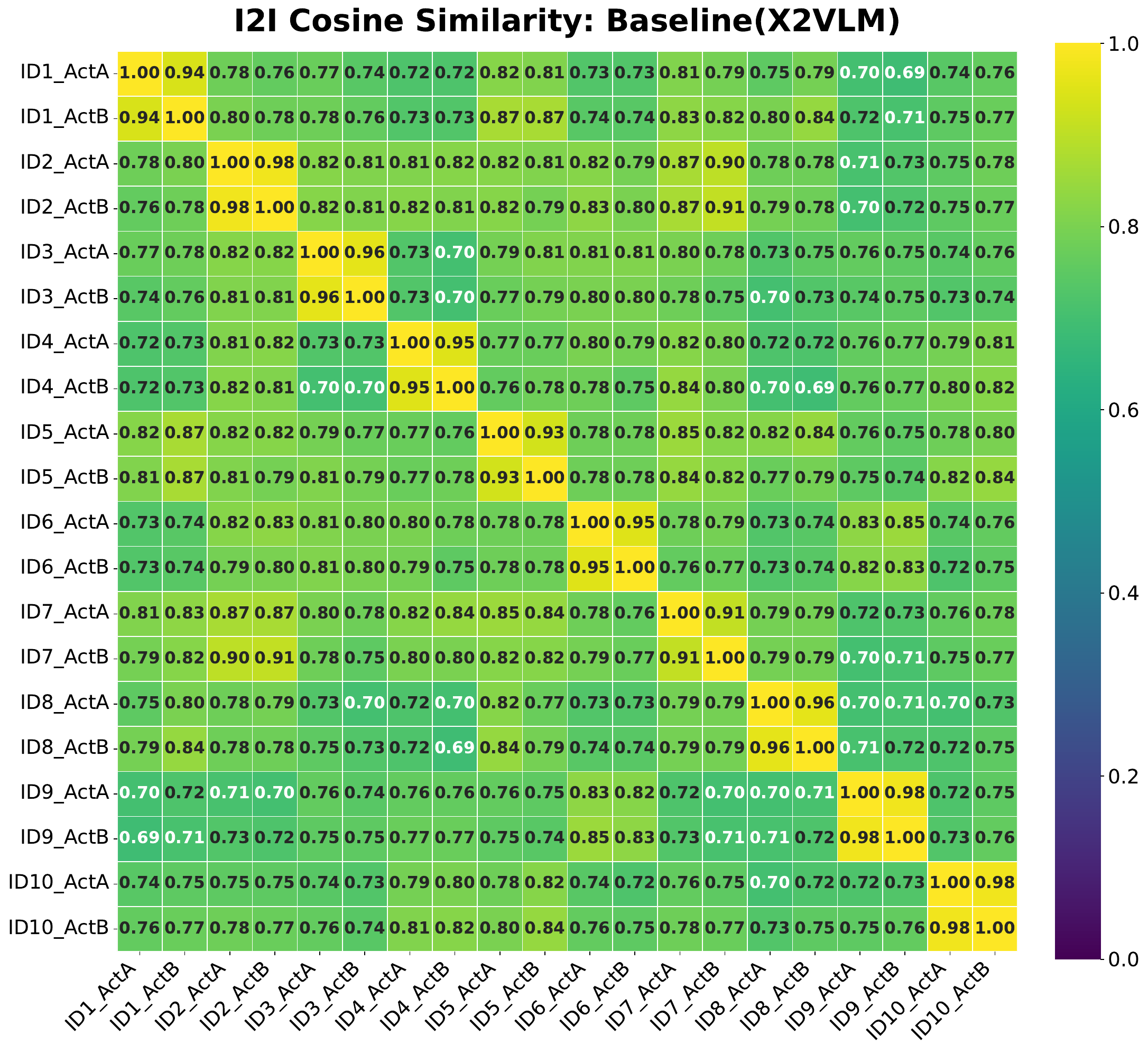}
        \caption{Image similarity of the baseline X2VLM on ten pairs of positive/hard negative samples.}
        \label{fig:I2I_Cos_Sim_X2VLM}
    \end{subfigure}
    \hfill
    \begin{subfigure}[b]{0.48\textwidth}
        \centering
        \includegraphics[width=\linewidth]{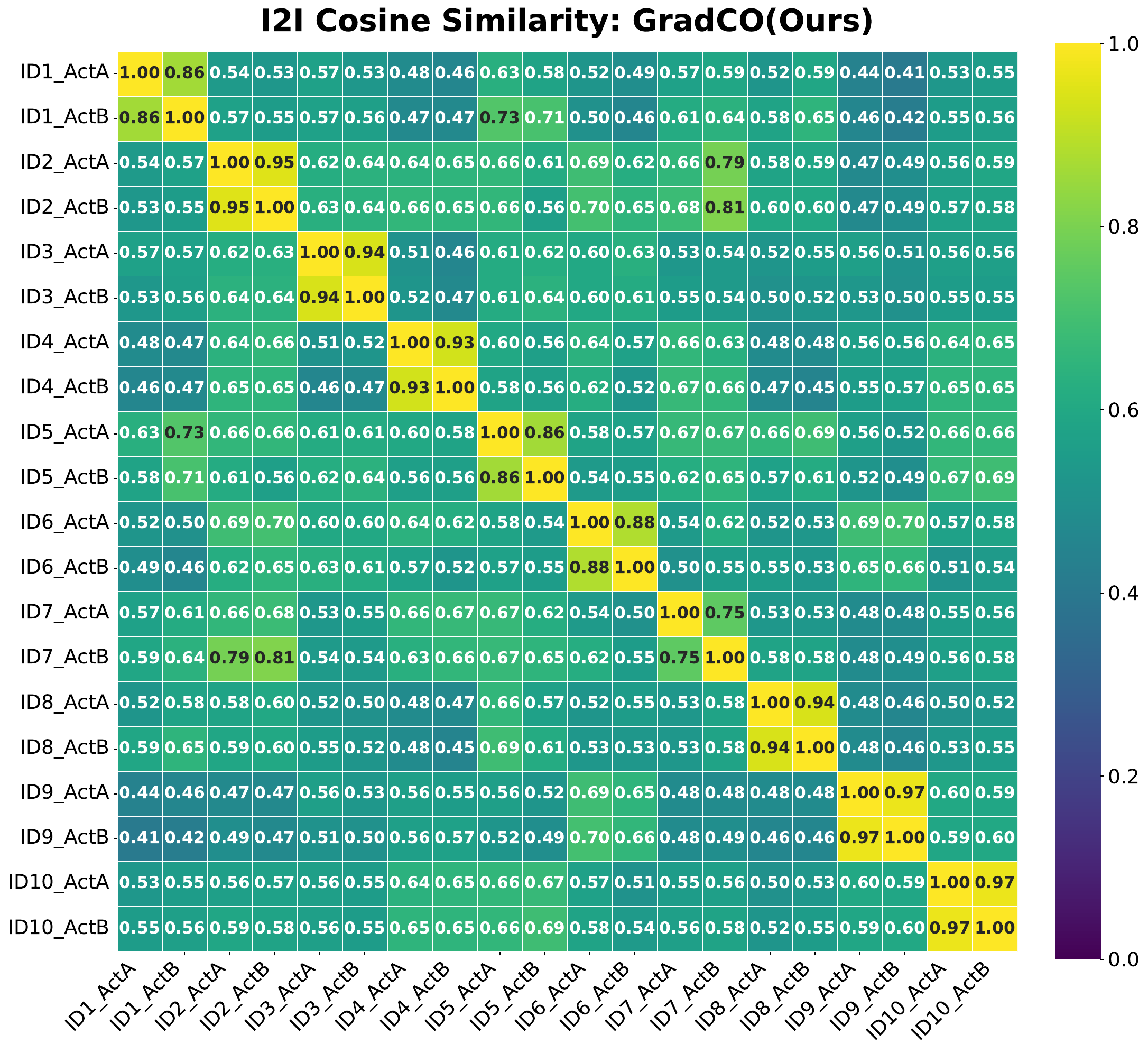}
        \caption{Image similarity of our GradCO model on ten pairs of positive/hard negative samples.}
        \label{fig:I2I_Cos_Sim_GradCO}
    \end{subfigure}
    \caption{\textbf{Visualization of hard-negative feature separation.}
Image-to-image cosine similarity matrices of ten selected hard-negative pairs
from the PAB test set. 
(a) X2VLM baseline shows a relatively diffuse similarity pattern with higher
cross-identity similarity.
(b) GradCO produces a clearer block-like structure, maintaining higher
within-identity similarity while reducing similarity among different
identities. 
The visualization qualitatively illustrates improved feature organization
under identity--behavior conflicts.
}
    \label{fig:I2I_Cos_Sim}
\end{figure*}

\begin{figure}[ht]
  \centering
  \includegraphics[width=\linewidth]{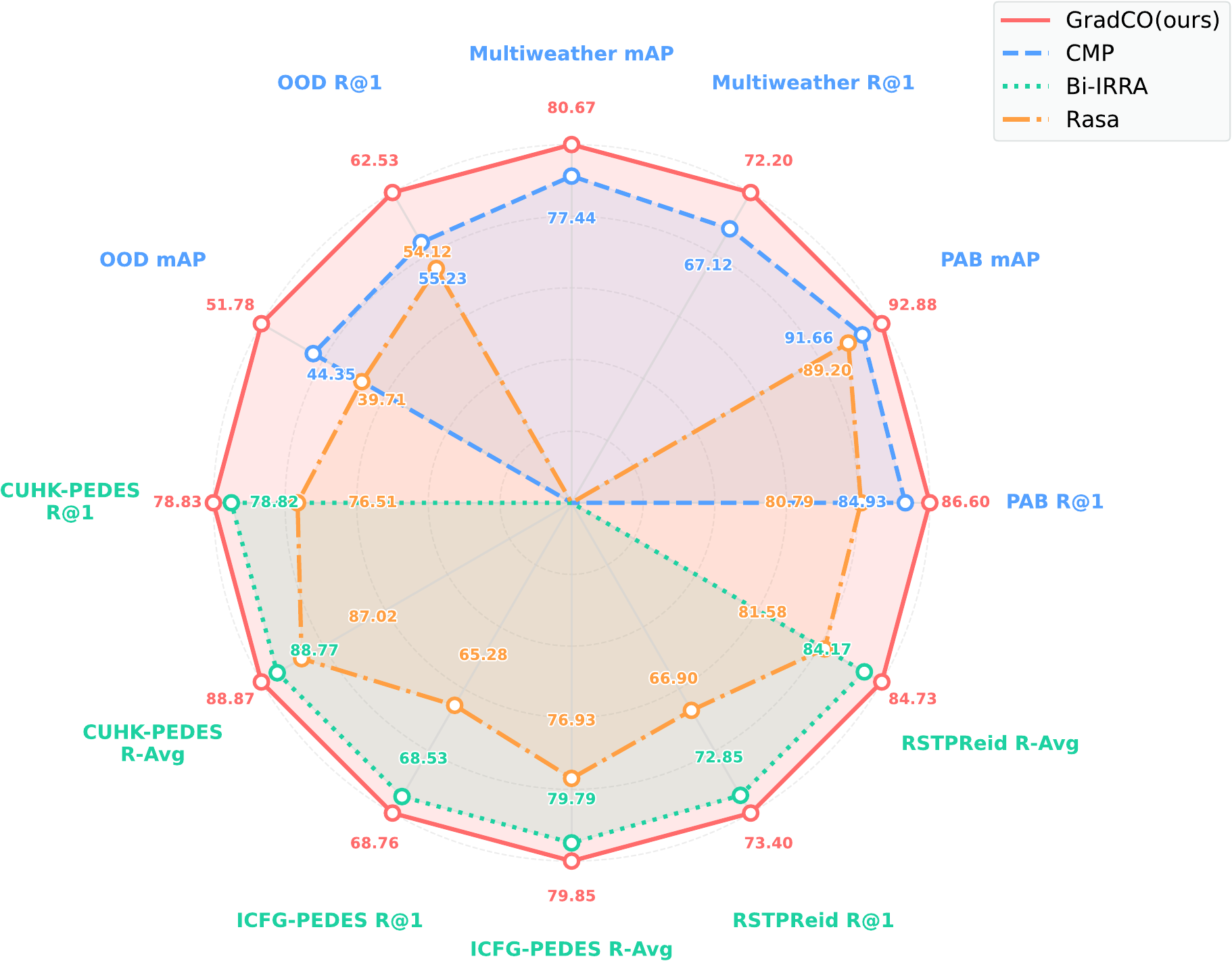}
  \caption{
  Radar comparison across the reported TPAS and TIPR metrics.
  GradCO achieves the strongest performance on the displayed TPAS metrics
  and obtains competitive results on TIPR benchmarks.
  }
  \label{fig:radar_chart}
\end{figure}

\begin{figure*}[t]
  \centering
   \resizebox{\linewidth}{!}{\includegraphics[width=1\linewidth]{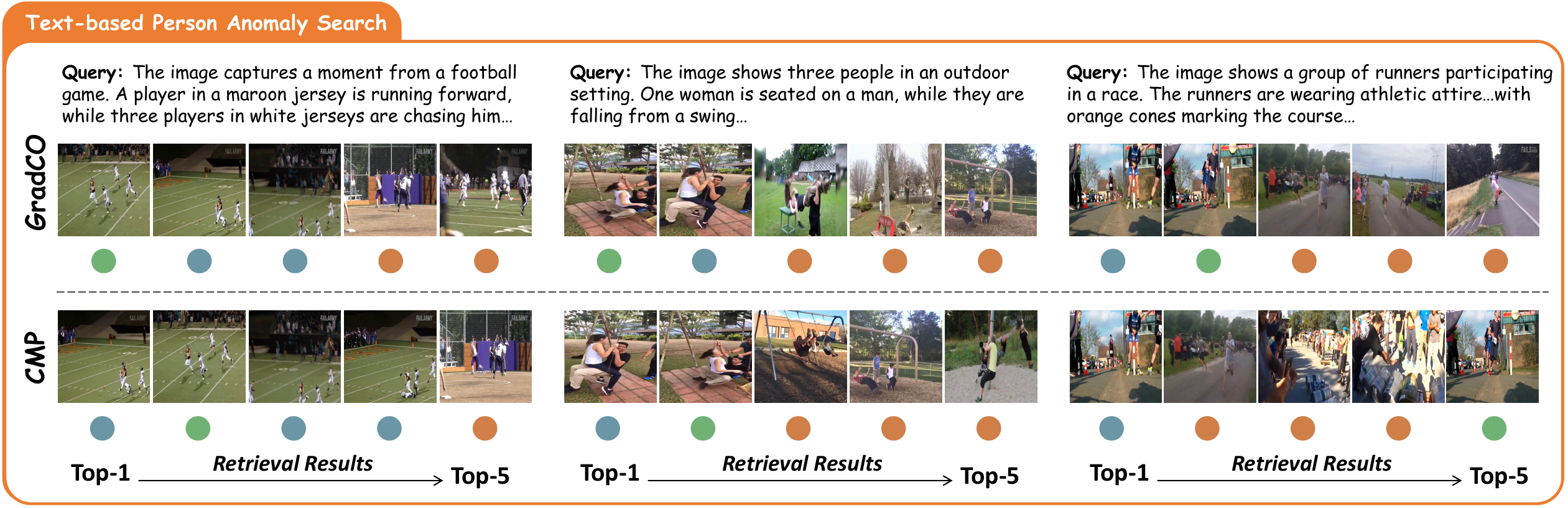}}
   \caption{Case study on PAB. Green dots \textcolor{greendot}{\scalebox{1.5}{\textbullet}} represent complete matches, blue dots \textcolor{bluedot}{\scalebox{1.5}{\textbullet}} represent identical IDs but mismatched behavior, and orange dots \textcolor{orangedot}{\scalebox{1.5}{\textbullet}} represent complete mismatches.}
   \label{fig:pab_case}
\end{figure*}

\begin{figure*}[!t]
  \centering
  \resizebox{\linewidth}{20cm}{\includegraphics[width=1\linewidth]{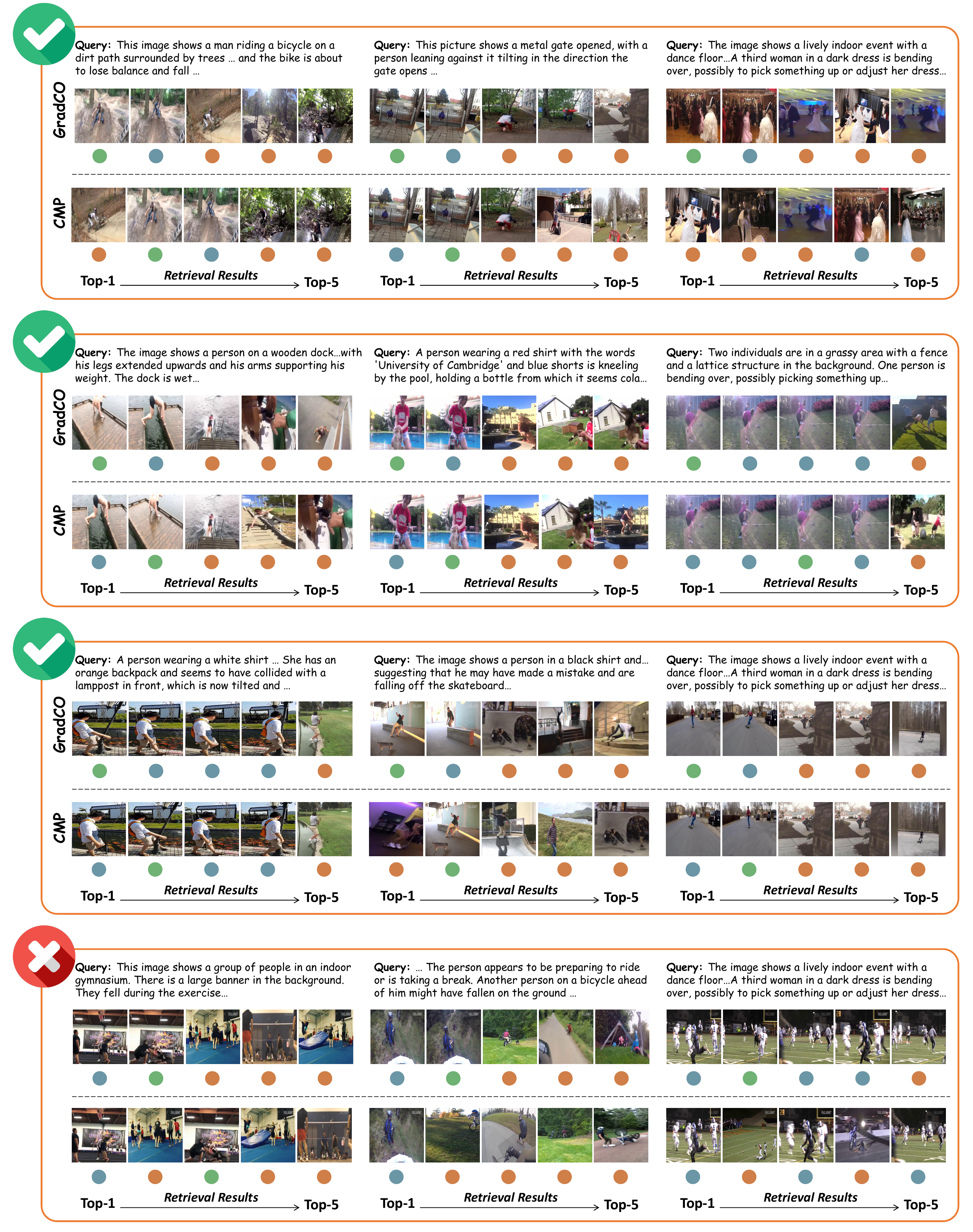}}
  \caption{
  \textbf{Qualitative retrieval comparison on PAB.}
  We select twelve representative TPAS cases involving challenging
  identity--behavior variations. Green dots
  \textcolor{greendot}{\scalebox{1.5}{\textbullet}} indicate complete matches,
  blue dots \textcolor{bluedot}{\scalebox{1.5}{\textbullet}} indicate
  identical identities with mismatched behaviors, and orange dots
  \textcolor{orangedot}{\scalebox{1.5}{\textbullet}} indicate complete
  mismatches.
  }
  \label{fig:Case}
\end{figure*}

\subsubsection{Visualization of Hard-negative Feature Separation}
\label{sup:qualitative_sim_vis}

To qualitatively analyze the representation structure under challenging
identity--behavior conflicts, we visualize the image-to-image cosine similarity
matrices of ten selected hard-negative pairs from the PAB test set
\cite{CMP2025}. The comparison between the X2VLM baseline and GradCO is shown
in Figure~\ref{fig:I2I_Cos_Sim}. Both models are evaluated on the same selected
samples, and the visualization provides an intuitive comparison of their
learned feature organization. Note that this qualitative analysis does not
isolate the individual effects of B-MoE and COL, but illustrates the overall
representation changes introduced by the proposed framework.

As shown in Figure~\ref{fig:I2I_Cos_Sim_X2VLM}, the X2VLM baseline exhibits a
relatively diffuse similarity pattern. Several off-diagonal entries between
different identities remain highly similar, with values mainly distributed in
the range of 0.70--0.80. This indicates that hard-negative samples with
similar appearance or behavior-related patterns remain difficult to separate
in the learned feature space, leading to ambiguous identity-level
representation.

In contrast, Figure~\ref{fig:I2I_Cos_Sim_GradCO} shows that GradCO produces a
clearer block-like similarity structure. Samples belonging to the same
identity maintain higher similarity, while samples from different identities
show reduced similarity. For example, the similarity between
$ID1\_ActA$ and $ID1\_ActB$ remains high (0.86), whereas some cross-identity
pairs, such as $ID1$ and $ID9$, are reduced to 0.44. These observations
suggest that GradCO learns a more discriminative organization of
identity-related and behavior-related information, improving the separation
of confusing hard-negative samples.

\subsubsection{Multi-dimensional Performance Comparison}
\label{sup:task_radar_chart}

As shown in Figure~\ref{fig:radar_chart}, the radar chart summarizes the reported TPAS and TIPR benchmark results to
provide an overall view of GradCO's performance across different evaluation
settings. Unlike individual metric comparisons, this visualization highlights
the relative performance trends among different methods, while the detailed
numerical comparisons are provided in the main tables.

For TPAS benchmarks, GradCO achieves the strongest performance on the
displayed PAB, Multi-Weather, and OOD metrics. In particular, the results on
Multi-Weather and UCC indicate improved performance under the tested
environmental variations and OOD conditions. These observations demonstrate
the effectiveness of GradCO on the evaluated TPAS scenarios, while they do not
imply generalization to all possible domain shifts.

For the UCC OOD evaluation, GradCO improves over CMP (1M) by 7.30 and 7.66
percentage points in R@1 and mAP, respectively. These improvements provide
empirical evidence of stronger retrieval performance on the tested OOD
benchmark. The contribution of individual components is further analyzed
through the corresponding ablation studies.

For TIPR benchmarks, GradCO remains competitive across the reported datasets.
As summarized in the radar chart and quantitative tables, GradCO achieves the
best displayed averages on CUHK-PEDES, ICFG-PEDES, and RSTPReID, leads UFine3C,
while Bi-IRRA remains higher on UFine6926 R-Avg. These results indicate that
the proposed framework maintains strong cross-modal retrieval capability
beyond the TPAS setting.

\subsection{Case Study}
\label{sup:qualitative_case}

We compare GradCO with the representative baseline CMP on representative
queries from the PAB dataset. Figure~\ref{fig:pab_case} presents three
queries, followed by twelve additional cases in Figure~\ref{fig:Case}.
These examples illustrate retrieval under identity--behavior ambiguity,
where visually similar candidates may exhibit different behaviors.

\noindent \textbf{Results in Figure~\ref{fig:pab_case}.}
From left to right, the three queries describe a woman and a man falling
from a swing, runners participating in a race, and a football game. GradCO
ranks the target first in the first two cases. In the swing example, the
query combines appearance cues, such as ``woman sitting on a man'' and
``swing'', with the behavior cue ``falling down''. CMP ranks a visually
similar image with mismatched behavior first, whereas GradCO retrieves the
target at Rank-1. In the third case, GradCO ranks a highly similar hard
negative first and the target second, while CMP ranks the target fifth.

\noindent \textbf{Additional successful cases in Figure~\ref{fig:Case}.}
Reading Figure~\ref{fig:Case} from left to right and from top to bottom,
the first three rows show nine further cases in which GradCO ranks the
target at Rank-1. The first row contains the imbalanced-cycling,
tilting-gate, and dance-floor cases. In the tilting-gate example, CMP
retrieves the same person standing upright in front of the gate, whereas
GradCO retrieves the target leaning against it. The second row contains
the handstand-at-the-pier, kneeling-with-a-bottle, and bending-over-on-
grass cases. In the handstand example, CMP retrieves the correct scene
and person but mismatches the behavior, while GradCO ranks the handstand
target first. The third row contains the lamppost-collision,
skateboard-fall, and indoor-bending cases. In the lamppost example,
GradCO distinguishes the collision moment from a visually similar
walking case.

\noindent \textbf{Failure cases.}
The final row of Figure~\ref{fig:Case} shows three further queries,
including the gymnasium example. In that example, GradCO ranks a
same-identity but behavior-mismatched sample first and the target second,
while CMP ranks the target fifth or lower. Thus, in both the football
case in Figure~\ref{fig:pab_case} and the gymnasium case in
Figure~\ref{fig:Case}, GradCO misses Rank-1 but ranks the target above
CMP. These cases show that GradCO can provide a more discriminative
ordering among difficult candidates, while also indicating that
behavior retrieval remains challenging in the presence of highly similar
candidates, visual clutter, or complex multi-person interactions.
\section{Conclusion}
In this paper, we investigate the TPAS task and reveal two key challenges of existing decoupling paradigms: gradient conflict and aggregation distortion. To address these challenges, we propose GradCO, a hierarchical Mixture-of-Experts framework that decouples appearance and behavior, alleviates hard-negative-induced optimization conflicts through Conflict Offload Learning, and enables intent-aware feature aggregation with Hierarchical Intent Alignment. Extensive experiments on five TPAS/TIPR benchmark datasets validate the effectiveness and robustness of our approach.

\bibliography{main}

\clearpage

\end{document}